\documentclass[10pt,journal,compsoc]{IEEEtran}
\usepackage{times}
\usepackage{epsfig}
\usepackage{graphicx}
\usepackage{amsmath}
\usepackage{amssymb}
\usepackage{multirow}
\usepackage{booktabs}
\usepackage{tabu}
\usepackage[dvipsnames]{xcolor}
\usepackage{spreadtab}
\usepackage{tabularx}

\DeclareMathOperator*{\argmin}{arg\,min}
\newcommand{\etal}{et al.}

\ifCLASSOPTIONcompsoc
  \usepackage[nocompress]{cite}
\else
  \usepackage{cite}
\fi
\ifCLASSINFOpdf
\else
\fi
\begin{document}
%
\title{Understanding Pixel-level 2D Image Semantics with 3D Keypoint Knowledge Engine}
%
%
%
%

\author{Yang You, Chengkun Li, Yujing Lou, Zhoujun Cheng, 
Liangwei Li, \\
Lizhuang Ma, Weiming Wang, Cewu Lu~\IEEEmembership{Member,~IEEE}
\IEEEcompsocitemizethanks{\IEEEcompsocthanksitem Yang You, Chengkun Li, Yujing Lou, Zhoujun Cheng, 
Liangwei Li, Lizhuang Ma, Weiming Wang and Cewu Lu are with Shanghai Jiao Tong University, Shanghai, 200240, China. Cewu Lu is also the member of Qing Yuan Research Institute, Shanghai Qizhi Research Institute and MoE Key Lab of Artificial Intelligence, AI Institute, Shanghai Jiao Tong University, China.
\\
\IEEEcompsocthanksitem Weiming Wang and Cewu Lu are the corresponding authors.\protect\\
E-mails: wangweiming@sjtu.edu.cn, lucewu@sjtu.edu.cn.
}
}

%
%

\markboth{Journal of \LaTeX\ Class Files,~Vol.~14, No.~8, August~2015}%
{Shell \MakeLowercase{\textit{et al.}}: Bare Demo of IEEEtran.cls for Computer Society Journals}
%



\IEEEtitleabstractindextext{%
\begin{abstract}
Pixel-level 2D object semantic understanding is an important topic in computer vision and could help machine deeply understand objects (e.g. functionality and affordance) in our daily life. However, most previous methods directly train on correspondences in 2D images, which is end-to-end but loses plenty of information in 3D spaces. In this paper, we propose a new method on predicting image corresponding semantics in 3D domain and then projecting them back onto 2D images to achieve pixel-level understanding. In order to obtain reliable 3D semantic labels that are absent in current image datasets, we build a large scale keypoint knowledge engine called KeypointNet, which contains 103,450 keypoints and  8,234 3D models from 16 object categories.  Our method leverages the advantages in 3D vision and can explicitly reason about objects self-occlusion and visibility. We show that our method gives comparative and even superior results on standard semantic benchmarks.
\end{abstract}

}

\maketitle

\IEEEdisplaynontitleabstractindextext

%
\IEEEpeerreviewmaketitle

\IEEEraisesectionheading{\section{Introduction}\label{sec:introduction}}

%
%
%
%
\IEEEPARstart{G}eneral object pixel-level semantic understanding is an important though tough topic in computer vision. Currently, plenty of deep convolutional methods have been applied to 2D images and gain much attention~\cite{chen2018encoder,seo2018attentive,ghiasi2019fpn,dai2016instance,he2017mask}. Among them, quite a few methods try to solve the problem of 2D semantic correspondence~\cite{choy2016universal,han2017scnet,kim2017fcss,horn1981determining,liu2010sift,rocco2017convolutional,rocco2018end,seo2018attentive,ham2017proposal,min2019dynamic,li2020correspondence} to achieve pixel-level understanding. That is, given a pair of images with different object but within the same category (such as two photos of different chairs taken from different angles and environments), find their semantically corresponding parts in the images. In this way, we can understand object deeply, such as how to grasp, manipulate and semantic editing. Note such kind of pixel-level understanding is largely different with segmentation problem that label pixel category only.  \cite{choy2016universal,han2017scnet,kim2017fcss,horn1981determining,liu2010sift} propose to match local regions between pairs of images and \cite{rocco2017convolutional,rocco2018end,seo2018attentive,ham2017proposal} formulate it as a global image alignment optimization problem.

However, these 2D CNN based methods can not reason about the internal 3D structure of objects in the image, making them vulnerable on view changes. Besides, previous 2D-based methods require a reference image to infer dense correspondences. These image references often experience severe occlusion and viewpoint changes, thus not suitable for pixel-wise semantic correspondence inference. 

\begin{figure}[ht]
    \centering
    \includegraphics[width=0.9\linewidth]{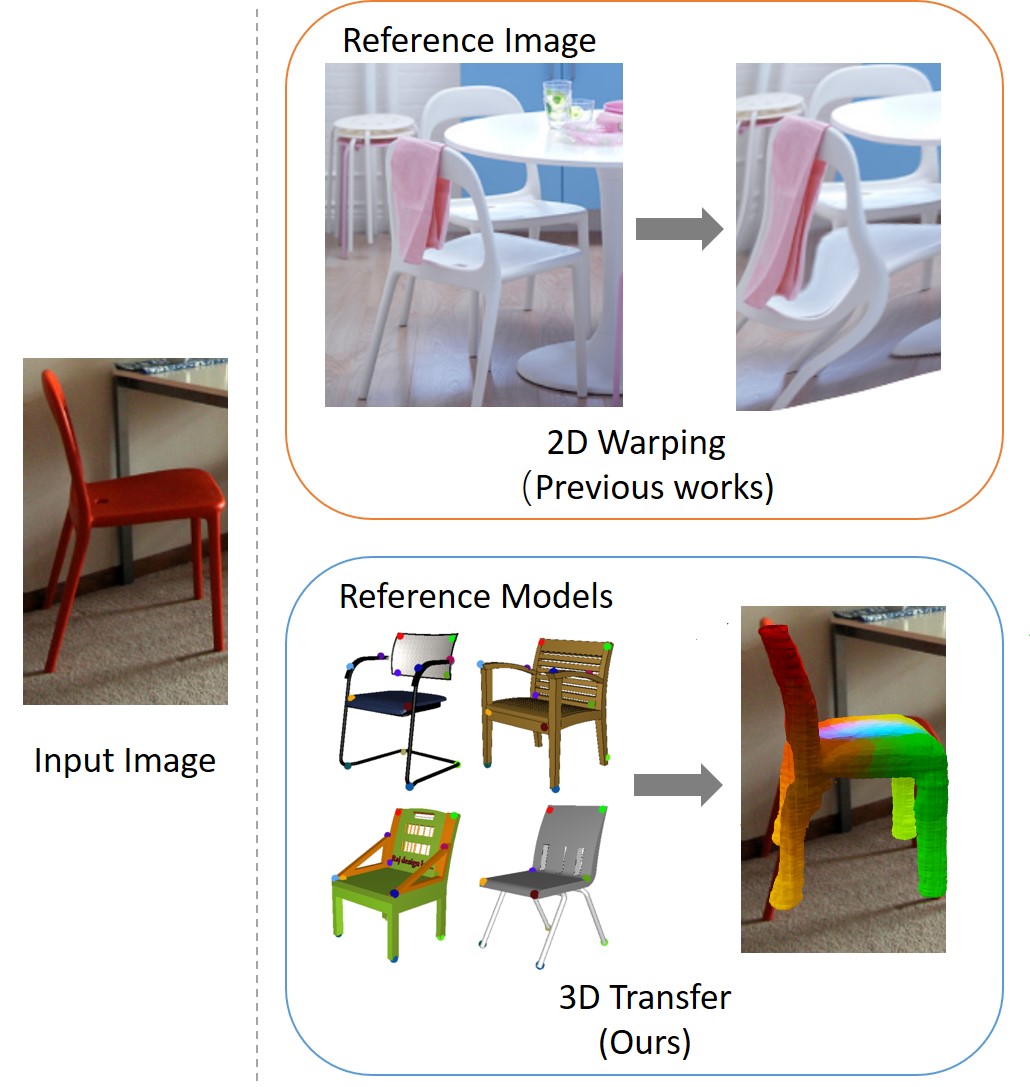}
    \caption{\textbf{Comparison between previous works and our method on 2D pixel-wise semantic understanding.} Previous methods tend to find pixel-wise semantics by transferring from a reference image with 2D warping. Viewpoint and occlusion variations make them fail to align well with the input image. Our method directly estimates viewpoints and transfer dense semantics from an existing large-scale 3D model repository.}
    \label{fig:intro}
\end{figure}



To overcome these problems, we doubt if one could directly define canonical semantic embeddings on 3D objects and then transfer these semantic embeddings from 3D space to 2D images. That is, we first estimate 3D structures from a single image and then feed such 3D structures into a semantic prediction module to get dense embeddings on 3D objects; finally we project the embeddings back onto the image. One of the key components is the 3D semantic prediction module. It accepts a general 3D object and outputs dense semantic embeddings for each point on the object. Thus, we know semantic on each pixel. 

It is natural to consider training a neural network based on some 3D object semantic knowledge engine. However, since we require fine-grained semantic embeddings, the knowledge engine should include dense semantic embeddings that are intra-class consistent on general 3D objects. Such a database is impractical since annotating every point's embedding by hand is impossible unless we approximate dense embeddings with sparse annotations.

Annotating sparse 3D keypoints is an amenable way to define object semantics and it is essential in many applications such as object matching, object tracking, shape retrieval and registration~\cite{mian2006three, bueno2016detection, wang2018learning}. Provided 3D keypoints, we would like to leverage contrastive loss to train a embedding prediction network that maximize the difference between distances on intra-class and inter-class keypoint pairs. It would give a reasonable approximation on dense embeddings since neural networks often approximate continuous functions and could infer those ``missing'' semantics outside keypoints.

\begin{figure*}[ht]
    \centering
    \includegraphics[width=0.9\linewidth]{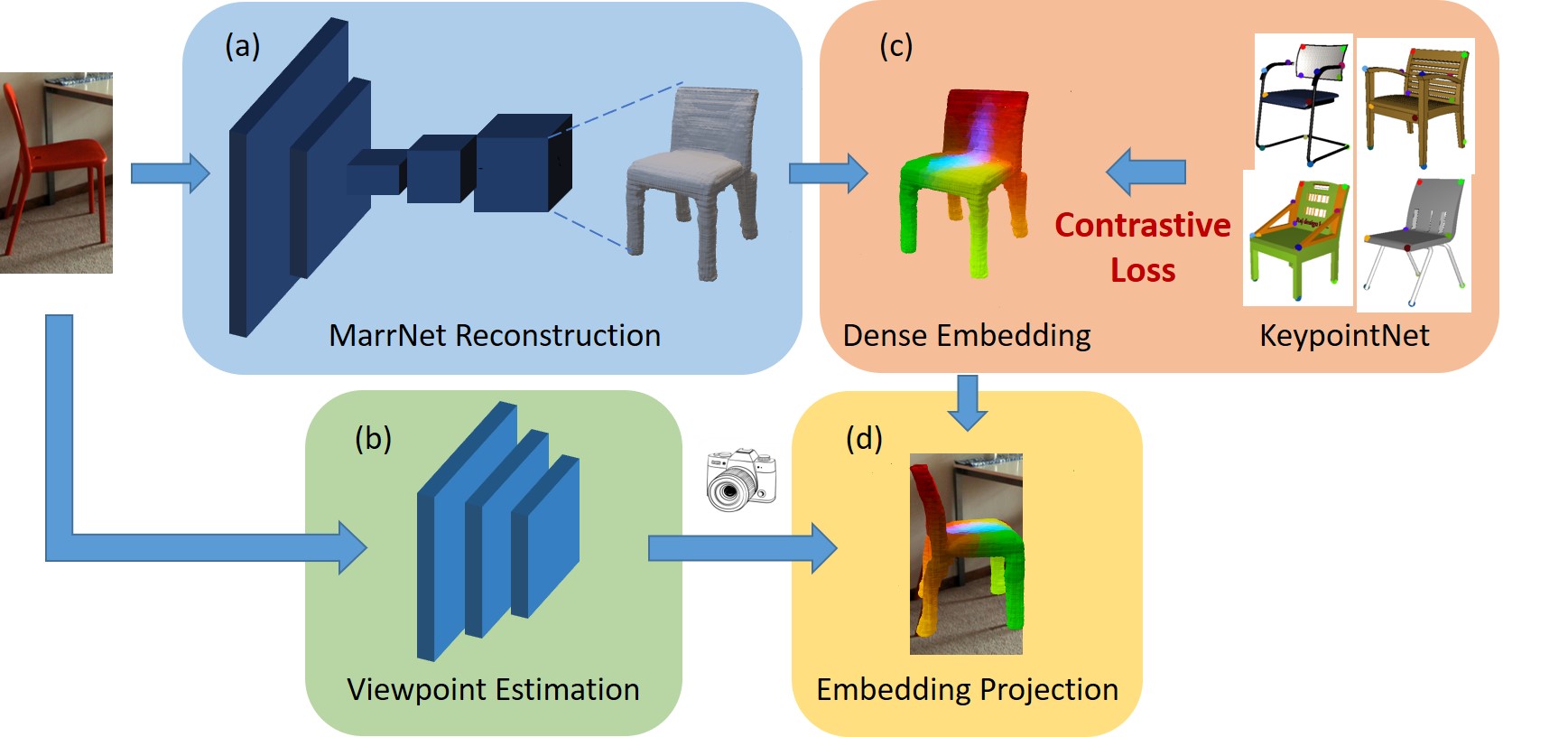}
    \caption{\textbf{Pixel-level semantic understanding pipeline}. (a) 3D models are reconstructed from single RGB images. (b) Viewpoints are also estimated from RGB images. (c) We obtain point-wise embeddings by training on existing 3D keypoint database, with contrastive loss. (d) Given viewpoints, 3D models and transferred embeddings, we project them onto the original image plane. In this way, we can achieve pixel-level semantic understanding.}
    \label{fig:app}
\end{figure*}

Currently, there are few 3D datasets focusing on the keypoint representation of an object. Dutagaci et al.~\cite{dutagaci2012evaluation} collect 43 models and label them according to annotations from various persons. Annotations from different persons are finally aggregated by geodesic clustering. ShapeNetCore keypoint dataset~\cite{yi2017syncspeccnn}, and a similar dataset~\cite{kim2013learning}, in another way, resort to an expert's annotation on keypoints, making them vulnerable and biased.
Therefore, in order to alleviate the bias of experts' definitions on keypoints, we ask a large group of people to annotate various keypoints according to their own understanding. Challenges rise in that different people may annotate different keypoints and we need to identify the consensus and patterns in these annotations. We propose a novel method to aggregate a large number of keypoint annotations from distinct people, by optimizing a fidelity loss. After this auto aggregation process, these generated keypoints are verified based on some simple priors such as symmetry. To this end, we build a large-scale and diverse keypoint knowledge engine named \textbf{KeypointNet} which contains 8,234 models with 103,450 keypoints. These keypoints are of high fidelity and rich in structural or semantic meanings. 
Some examples are given in Figure~\ref{fig:vis}. Given sparse keypoint annotations, we could approximate dense 3D embeddings by training a pointwise neural network with contrastive loss, where keypoints with same semantics form positive pairs while negative pairs come from different keypoint semantics.

In the previous paragraph, we describe how the 3D knowledge database is built and how 3D semantic prediction module is made. Recall that we first estimate 3D structures from a single image and then  feed  such  3D  structures  into  a  semantic  prediction  module to  get  dense  embeddings  on  3D  objects;  finally  we  project  the embeddings back onto the image. In order to estimate 3D structures from a single image, we gain insights from ShapeHD~\cite{wu2018learning} and build a network based on it. It first estimates 2.5D sketches including depths, normals and silhouettes from an RGB image, and then predicts the 3D shape in canonical pose from 2.5D sketches. To project dense embeddings back onto 2D image, one has to know exact camera parameters to do so. We leverage a hierarchical prediction approach, where a coarse camera pose is estimated by a 2D convolutional network and then a differentiable renderer is optimized to match predicted and projected silhouettes. The full pipeline is shown in Figure~\ref{fig:app}.

\begin{table*}[ht]
\small
\begin{minipage}{\textwidth}

\centering
\begin{tabular}{l|c|c|c|c|c|c|c}
\hline
\textbf{Dataset}  & Domain & Correspondence &Template-free & Instances & Categories & Keypoints & Format\\
\hline
\textbf{Bogo \etal}~\cite{Bogo:CVPR:2014}  &human &  $\surd$& $\times$&100 & 1& 689K& mesh\\
\textbf{Yi \etal}~\cite{yi2017syncspeccnn}&chair &  $\surd$& $\times$ &6243 & 1& $\sim$60K& point cloud\\
\textbf{Dutagaci \etal}~\cite{dutagaci2012evaluation} &general&  $\times$ & $\surd$ & 43& 16& $<$1K &mesh\\
\textbf{Kim \etal}~\cite{kim2012exploring}& general& $\surd$& $\times$& 404 & 4& $\sim$3K& mesh\\
\textbf{Xiang \etal}~\cite{xiang_wacv14} &general & $\times$& $\times$& 36292 & 12& 150K+&RGB w. 3D model\\
\hline
\textbf{Ours}&general& $\surd$& $\surd$& 8234 & 16& 103K+&point cloud \& mesh\\
\hline
\end{tabular}
\bigbreak
\caption{\textbf{Comparison of 3d keypoint datasets}. \textbf{Correspondence} indicates whether keypoints are indexed correspondingly. \textbf{Template-free} indicates whether it avoids hardcoded keypoint templates.}
\label{tab:comparison}
\end{minipage}
\end{table*}

In summary, we make the following contributions:
\begin{itemize}
    \item To the best of our knowledge, we build the first large-scale knowledge engine on 3D keypoints, both in number of categories and keypoints to achieve pixel-level understanding and semantic transfer.
    
    \item We come up with a novel approach on aggregating people's annotations on keypoints, even if their annotations are independent from each other.
    
    \item We propose a novel pipeline trying to solve general 2D object semantic understanding by leveraging it to the 3D domain.
\end{itemize}

Our paper is organized as follows: in Section 2, we briefly discuss related works, involving various aspects; in Section 3, we introduce our keypoint knowledge engine and how it is built; in Section 4, we describe how we utilize the knowledge engine to predict 2D dense embeddings, together with various experiment results.

A preliminary version of this work was presented in CVPR 2020 \cite{you2020keypointnet}. In this study, we extend it substantially by proposing a new pipeline on 2D general object semantic understanding based on our 3D keypoint engine. The whole section~\ref{sec:pipeline} describes our novel pipeline in detail and conducts various experiments to validate our design. Results show that our method outperforms previous state-of-the-art methods on various 2D object semantic understanding benchmarks.

\section{Related Work}

\subsection{2D Object Semantic Understanding}
Previous works on 2D object semantic understanding can be generally classified into two subjects: one is to directly predict pixel-wise embeddings with an end-to-end network while the other is to transfer dense semantics from an existing reference image.

For the first subject, there are plenty of works~\cite{porzi2019seamless,chen2017deeplab} on estimating a semantic label for each pixel. This task is also named semantic segmentation. Besides, some instance segmentation methods~\cite{girshick2014rich,girshick2015fast,ren2015faster,redmon2016you} also try to give every pixel an instance ID with semantics. Afterwards, panoptic segmentation~\cite{kirillov2019panoptic} is proposed to combine both semantic and instance segmentation. However, all these segmentation methods lack find-grained semantics, such as ``which pixel is on the arm of chair'', ``where is the nose of aeroplane'', etc.

For the second subject, which is also known as dense image semantic correspondence, there is a long history that dates back to optical flow~\cite{horn1981determining} and multi-stereo~\cite{okutomi1993multiple} based methods. Recently, some local descriptor based methods~\cite{ham2017proposal,liu2010sift}  are explored to find dense correspondences across different objects. With the advance of deep learning, neural features~\cite{hariharan2015hypercolumns,lin2017feature,kong2016hypernet} are broadly used as they are more robust and generalizable. Methods like Seo \etal~\cite{seo2018attentive}, Rocco \etal~\cite{rocco2018neighbourhood} and Min \etal~\cite{min2019hyperpixel} view semantic correspondence as a matching problem in high-dimensional feature images. In addition, Florence \etal~\cite{florence2018dense} and Schmidt \etal~\cite{schmidt2016self} leverage an unsupervised methods to learn consistent dense embeddings with SLAM across different objects. Guler \etal~\cite{guler2018densepose} learn dense correspondences between human poses by first training a ``teacher'' network where loss is defined on sparse keypoint annotations and then ``inpainting'' dense values across the full image domain. However, their method only considers 2D image domain inpainting, ignoring the underlying 3D structure and viewpoint changes. Min \etal~\cite{min2019dynamic} introduce a novel approach to visual correspondence that dynamically composes effective features by leveraging relevant
layers conditioned on the images to match. Li \etal~\cite{li2020correspondence} propose non-isotropic 4D convolution kernel and a simple and efficient multi-scale self-similarity module to realize one-to-one robust intra-class matching.

\subsection{3D Object Semantic Understanding}
Allen \etal~\cite{allen2003space} and Blanz \etal~\cite{blanz1999morphable} are the pioneers on detecting 3D semantic correspondence between human bodies and faces. Recently, some unsupervised methods~\cite{halimi2018self,roufosse2019unsupervised,groueix20183d} are proposed on learning dense correspondences between humans and animals. With the help of recent large scale model dataset such as Chang \etal~\cite{chang2015shapenet} and Mo \etal~\cite{mo2019partnet}, finding semantic correspondences on general objects become possible. Sung \etal~\cite{sung2018deep} and Yi \etal~\cite{yi2017syncspeccnn} all learn a set of synchronized base functions in order to obtain dense correspondence from functional maps. In addition to ShapeNet, \cite{pavlakos20176,kim2013learning,you2019fine,you2020keypointnet} provide additional keypoint or correspondence annotations for object semantic understandings. 

Perhaps, Kulkarni \etal~\cite{kulkarni2019canonical} and Zhou et al.~\cite{zhou2016learning} are the closest to this paper. However, they assume that for all images, there is a template 3D model that fits well, making them not directly applicable to categories where the shapes across instances differ significantly in topology or undergo large articulation. Besides, they implicitly infer 3D models by generating a 2D-3D pixel maps while we explicitly predict each image's corresponding 3D shape.

\subsection{Keypoint-based Knowledge Engine}
Detection of 3D keypoints has been a very important task for 3D object understanding which can be used in many applications, such as object pose estimation, reconstruction, matching, segmentation, etc. Researchers have proposed various methods to produce interest points on objects to help further objects processing. Traditional methods like 3D Harris~\cite{sipiran2011harris}, HKS~\cite{sun2009concise}, Salient Points~\cite{castellani2008sparse}, Mesh Saliency~\cite{lee2005mesh}, Scale Dependent Corners~\cite{novatnack2007scale}, CGF~\cite{khoury2017learning}, SHOT~\cite{tombari2010unique}, etc, exploit local reference frames (LRF) to extract geometric features as local descriptors. 
However, these methods only consider the local geometric information without semantic knowledge, which forms a gap between detection algorithms and human understanding.

Recently, deep learning methods Yi \etal~\cite{yi2017syncspeccnn}, Sung \etal~\cite{sung2018deep} are proposed to detect keypoints. Unlike traditional ones, these methods do not handle rotations well. Though some recent methods~\cite{cohen2018spherical,you2018prin} try to fix this, deep learning methods still rely on ground-truth keypoint labels annotated
by human with expert verification.

Keypoint datasets have its origin in 2D images, where plenty of datasets on human skeletons and object interest points are proposed. For human skeletons, MPII human pose~\cite{andriluka14cvpr}, MSCOCO keypoint challenge~\cite{mscoco} and PoseTrack~\cite{andriluka2018posetrack} annotate millions of keypoints on humans. For more general objects, SPair-71k~\cite{min2019spair} contains 70,958 image pairs with diverse variations in viewpoint and scale, with a number of corresponding keypoints on each image pair. CUB~\cite{WahCUB_200_2011} provides 15 part locations on 11,788 images from 200 bird categories and PF-PASCAL~\cite{ham2017proposal} provides keypoint annotations for 20 object categories. Besides, LVIS~\cite{gupta2019lvis} provides a large
dataset for large vocabulary instance segmentation, which collects 2.2 million high-quality instance segmentation masks for over 1000 entry-level object categories in 164k images.



Keypoint datasets on 3D objects, include Dutagaci \etal~\cite{dutagaci2012evaluation}, Yi \etal~\cite{yi2017syncspeccnn} and Kim \etal~\cite{kim2013learning}. Dutagaci \etal~\cite{dutagaci2012evaluation} aggregates multiple annotations from different people with an ad-hoc method while the dataset is extremely small. Though Yi \etal~\cite{yi2017syncspeccnn}, Pavlakos \etal \cite{pavlakos20176} and Kim \etal~\cite{kim2013learning} give a relatively large keypoint dataset, they rely on a manually designed template of keypoints, which is inevitably biased and flawed. The differences between theirs and ours are illustrated in Table~\ref{tab:comparison}.

\subsection{Single View Shape Reconstruction} Recently, many works have been introduced on single view shape reconstruction. For supervised methods where a ground-truth model is available, Fan \etal~\cite{fan2017point} and Lin \etal~\cite{lin2018learning} reconstruct point clouds from single-view RGB images. Yao \etal~\cite{yao2019front2back} predicts per-pixel depth, which is then converted into a point cloud.  \cite{wu2017marrnet,chen2009learning,wu2016learning} predict voxel grids with a relatively small resolution while some other methods~\cite{park2019deepsdf,mescheder2019occupancy,liu2019learning} reconstruct implicit surface functions, where resolutions are not limited compared to voxels. In addition, there are also plenty of researches~\cite{gkioxari2019mesh,wen2019pixel2mesh++} focused on triangle mesh reconstruction, which is constrained by mesh topology. Pan et al.~\cite{pan2019deep} tries to modify the mesh topology during reconstruction. What's more, some other directions like reconstructing images as geometric primitive collections~\cite{gao2019sdm,tian2019learning} and complex octree structures~\cite{riegler2017octnet,tatarchenko2017octree} are also explored. 

For unsupervised single view shape prediction, \cite{sitzmann2019deepvoxels,niemeyer2019differentiable,rematas2019neural,eslami2018neural} utilize only 2D image annotations, together with a multi-view consistency prior, to reconstruct the implicit 3D models. Other works~\cite{wang2020deep,li2019synthesizing} focus on a large collection of images in the wild and reconstruct a model for each distinguished image.

\subsection{Differentiable Renderer}
Differentiable renderer is an emerging topic in recent years, we see that ~\cite{sitzmann2019deepvoxels,niemeyer2019differentiable} all utilize differentiable projections to learn a 3D shape from its 2D image projections. Kato \etal~\cite{kato2018neural} first brings this idea to mesh rasterization rendering and Li \etal~\cite{li2018differentiable} comes up with differentiable monte-carlo ray tracing. Jiang \etal~\cite{jiang2019sdfdiff} renders implicit surfaces defined by signed distance function in a differentiable way.

\section{Building KeypointNet: A Large-scale 3D Keypoint Knowledge Engine}

\subsection{Data Collection}
KeypointNet is built on ShapeNetCore~\cite{chang2015shapenet}. ShapeNetCore covers 55 common object categories with about 51,300 unique 3D models. 

 We filter out those models that deviate from the majority and keep at most 1000 instances for each category in order to provide a balanced dataset. In addition, a consistent canonical orientation is established (e.g., upright and front) for every category because of the incomplete alignment in ShapeNetCore.


We let annotators determine which points are important, and same keypoint indices should indicate same meanings for each annotator. Though annotators are free to give their own keypoints, three general principles should be obeyed: (1) each keypoint should describe an object’s semantic information shared across instances of the same object category, (2) keypoints of an object category should spread over the whole object and (3) different keypoints have distinct semantic meanings. After that, we utilize a heuristic method to aggregate these points, which will be discussed in Section~\ref{sec:aggr}.


Keypoints are annotated on meshes and these annotated meshes are then downsampled to 2,048 points. Our final dataset is a collection of point clouds, with keypoint indices.

\subsection{Dataset Statistics}
\begin{figure*}[ht]
    \centering
    \includegraphics[width=\linewidth]{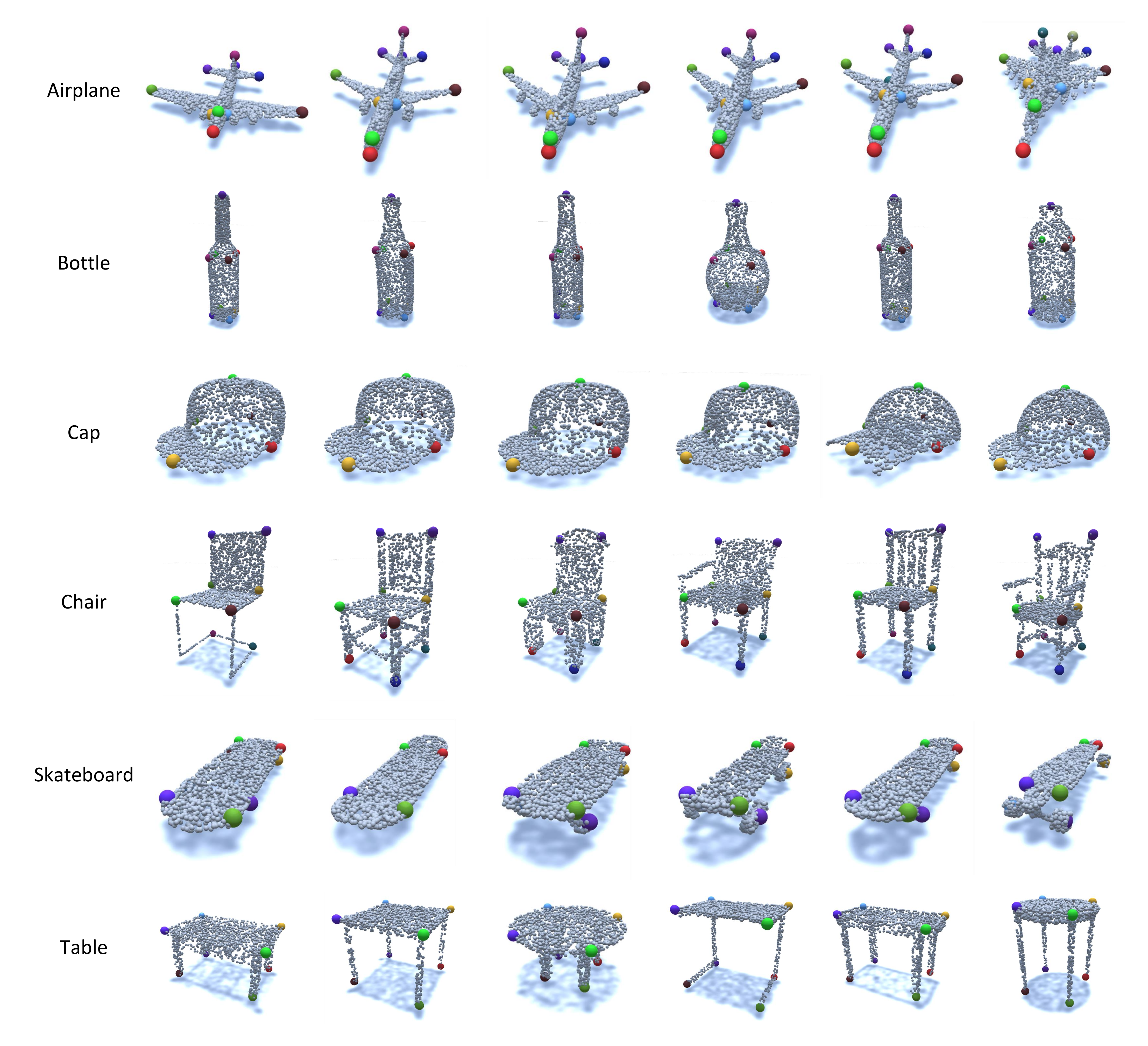}
    \caption{\textbf{Dataset Visualization.} Here we plot ground-truth keypoints for several categories. We can see that by utilizing our automatic aggregation method, keypoints of high fidelity are extracted.}
    \label{fig:vis}
\end{figure*}

At the time of this work, our dataset has collected 16 common categories from ShapeNetCore, with 8234 models. Each model contains 3 to 24 keypoints. Our dataset is divided into train, validation and test splits, with 7:1:2 ratio.  Table~\ref{tab:stat_kp} gives detailed statistics of the number of keypoints on each individual model while Table~\ref{tab:statistics} gives the total number of models and keypoints in each category. Some visualizations of our dataset is given in Figure~\ref{fig:vis}.

\begin{table}[ht]
\begin{minipage}{\linewidth}
\centering
\begin{tabular}{l|c|c|c}
\hline
\textbf{Category}  & \textbf{Minimum} & \textbf{Maximum} & \textbf{Median} \\
\hline
\textbf{Airplane} & 5 & 17 & 14\\
\textbf{Bathtub} & 8 & 24 & 16\\
\textbf{Bed} & 8 & 16 & 12\\
\textbf{Bottle} & 4 & 18 & 17\\
\textbf{Cap} & 5 & 6 & 6\\
\textbf{Car} & 14 & 22 & 22\\
\textbf{Chair} & 8 & 17 & 10\\
\textbf{Guitar} & 4 & 9 & 9 \\
\textbf{Helmet} & 5 & 9 & 9 \\
\textbf{Knife} & 4 & 6 & 6 \\
\textbf{Laptop} & 6 & 6 & 6\\
\textbf{Motorcycle} & 7 & 14 & 13\\
\textbf{Mug} & 10 & 11 & 11 \\
\textbf{Skateboard} & 6 & 10 & 10 \\
\textbf{Table} & 7 & 12 & 8\\
\textbf{Vessel} & 5 & 21 & 15 \\
\hline
\end{tabular}
\bigbreak
\caption{\textbf{The statistics of number of keypoints on each model, collected on the full dataset.} On each model, a sparse set of keypoints are labeled, ranging from 4 to 22.}
\label{tab:stat_kp}
\end{minipage}
\end{table}


\subsection{Annotation Tools}
\begin{figure}[ht]
    \centering
    \includegraphics[width=\linewidth]{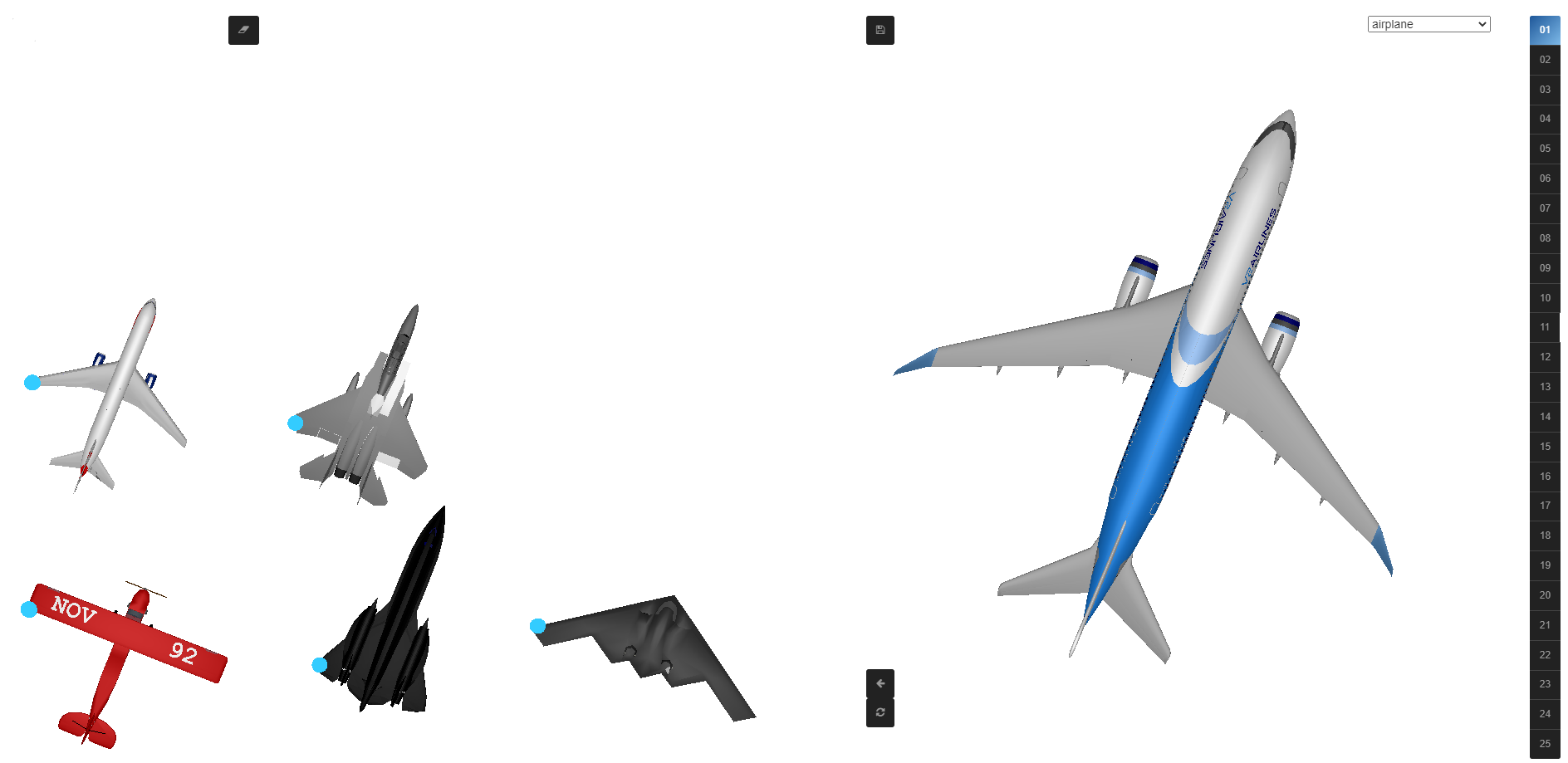}
    \caption{\textbf{Browser interface of the annotation tool.}}
    \label{fig:web}
\end{figure}

We develop an easy-to-use web annotation tool based on NodeJS. Every user is allowed to click up to 24 interest points according to his/her own understanding. The UI interface is shown in Figure~\ref{fig:web}.  Annotated models are shown in the left panel while the next unprocessed model is shown in the right panel.

\begin{figure*}[ht]
    \centering
    \includegraphics[width=\linewidth]{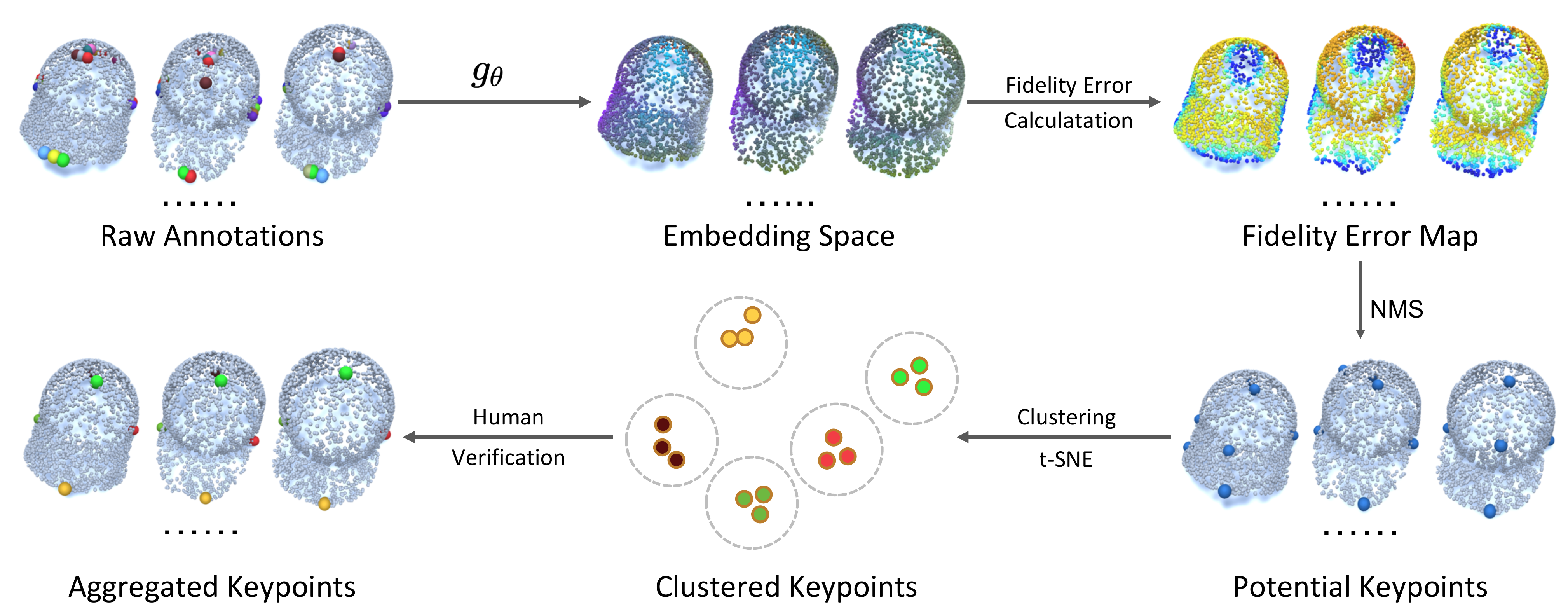}
    \caption{\textbf{Keypoint aggregation pipeline.} We first infer dense embeddings from human labeled raw annotations. Then fidelity error maps are calculated by summing embedding distances to human labeled keypoints. Non Minimum Suppression is conducted to form a potential set of keypoints. These keypoints are then projected onto 2D subspace with t-SNE and verified by humans.}
    \label{fig:pipeline}
\end{figure*}

\begin{table*}[ht]
\begin{minipage}[b]{.5\linewidth}
\begin{center}
\resizebox{!}{0.4\textwidth}{
\begin{spreadtab}{{tabular}{l|c|c|c|c|c}}
\hline
@\textbf{Category}  & @\textbf{Train} & @\textbf{Val} & @\textbf{Test} & @\textbf{All} & @\textbf{\#Annotators} \\
\hline
@\textbf{Airplane} & 715 & 102 & 205 & sum(b2:d2) & 26\\
@\textbf{Bathtub} & 344 & 49 & 99 & sum(b3:d3) & 15\\
@\textbf{Bed} & 102 & 14 & 30 & sum(b4:d4) & 8\\
@\textbf{Bottle} & 266 & 38 & 76 & sum(b5:d5) & 9\\
@\textbf{Cap} & 26 & 4 & 8 & sum(b6:d6) & 6\\
@\textbf{Car} & 701 & 100 & 201 & sum(b7:d7) & 17\\
@\textbf{Chair} & 699 & 100 & 200 & sum(b8:d8) & 15\\
@\textbf{Guitar} & 487 & 70 & 140 & sum(b9:d9) & 13\\
@\textbf{Helmet} & 62 & 10 & 18 & sum(b10:d10) & 8\\
@\textbf{Knife} & 189 & 27 & 54 & sum(b11:d11) & 5\\
@\textbf{Laptop} & 307 & 44 & 88 & sum(b12:d12) & 12\\
@\textbf{Motorcycle} & 208 & 30 & 60 & sum(b13:d13) & 7\\
@\textbf{Mug} & 130 & 18 & 38 & sum(b14:d14) & 9\\
@\textbf{Skateboard} & 98 & 14 & 29 & sum(b15:d15) & 9\\
@\textbf{Table} & 786 & 113 & 225 & sum(b16:d16) & 17\\
@\textbf{Vessel} & 637 & 91 & 182 & sum(b17:d17) & 18\\
\hline
@\textbf{Total} & sum(b2:b17) & sum(c2:c17) & sum(d2:d17) & sum(e2:e17) & sum(f2:f17)\\
\hline
\end{spreadtab}
}
\end{center}
\end{minipage}
\hfill
\begin{minipage}[b]{.5\linewidth}
\begin{center}
\resizebox{!}{0.4\textwidth}{
\begin{spreadtab}{{tabular}{l|c|c|c|c}}
\hline
@\textbf{Category}  & @\textbf{Train} & @\textbf{Val} & @\textbf{Test} & @\textbf{All} \\
\hline
@\textbf{Airplane} & 9695 & 1379 & 2756 & sum(b2:d2) \\
@\textbf{Bathtub} & 5519 & 772 & 1589 & sum(b3:d3) \\
@\textbf{Bed} & 1276 & 188 & 377 & sum(b4:d4) \\
@\textbf{Bottle} & 4366 & 625 & 1269 & sum(b5:d5) \\
@\textbf{Cap} & 154 & 24 & 48 & sum(b6:d6) \\
@\textbf{Car} & 14976 & 2133 & 4294 & sum(b7:d7) \\
@\textbf{Chair} & 8488 & 1180 & 2395 & sum(b8:d8) \\
@\textbf{Guitar} & 4112 & 591 & 1197 & sum(b9:d9) \\
@\textbf{Helmet} & 529 & 85 & 162 & sum(b10:d10) \\
@\textbf{Knife} & 946 & 136 & 276 & sum(b11:d11) \\
@\textbf{Laptop} & 1842 & 264 & 528 & sum(b12:d12) \\
@\textbf{Motorcycle} & 2690 & 394 & 794 & sum(b13:d13) \\
@\textbf{Mug} & 1427 & 198 & 418 & sum(b14:d14) \\
@\textbf{Skateboard} & 903 & 137 & 283 & sum(b15:d15) \\
@\textbf{Table} & 6325 & 913 & 1809 & sum(b16:d16) \\
@\textbf{Vessel} & 9168 & 1241 & 2579 & sum(b17:d17) \\
\hline
@\textbf{Total} & sum(b2:b17) & sum(c2:c17) & sum(d2:d17) & sum(e2:e17) \\
\hline
\end{spreadtab}
}
\end{center}
\end{minipage}
\bigbreak
\caption{\textbf{Keypoint Dataset statistics.} Left: total number of models in each category. Right: total number of keypoints in each category.}
\label{tab:statistics}
\end{table*}

\subsection{Keypoint Aggregation}
\label{sec:aggr}
Given all human labeled raw keypoints, we leverage a novel method to aggregate them together into a set of \textit{ground-truth} keypoints. 

There are generally two reasons: 1) distinct people may annotate different sets of keypoints and human labeled keypoints are sometimes erroneous, so we need an elegant way to aggregate these keypoints; 2) a simple clustering algorithm would fail to distinguish those closely spaced keypoints and cannot give consistent semantic labels.

\subsubsection{Problem Statement} 
Given a $2$-dimensional sub-manifold $\mathcal{M}_m\subset \mathbb{R}^{3}$, where $m$ is the index of the model, a valid annotation from the $c$-th person is a keypoint set $\{l_{m,k}^{(c)}|l_{m,k}^{(c)} \in\mathcal{M}_m\}_{k=1}^{K_c}$, where $k$ is the keypoint index and $K_c$ is the number of keypoints annotated by person $c$. Note that different people may have different sets of keypoint indices and these indices are independent.

Our goal is to aggregate a set of potential ground-truth keypoints $\mathcal{Y}=\{y_{m,k}|y_{m,k} \in \mathcal{M}_m, m=1,\dots M,k=1,\dots K_m \}$, where ${K_m}$ is the number of proposed keypoints for each model $\mathcal{M}_m$, so that $y_{m_1,k}$ and $y_{m_2,k}$ share the same semantic.

\subsubsection{Keypoint Saliency}
Each annotation is allowed to be erroneous within a small region, so that a keypoint distribution is defined as follows:
\begin{align*}
    p(x|x\text{ is the $k$-th keypoint}, x\in\mathcal{M}_m) = \frac{\phi(l_{m,k}, x)}{Z(\phi)},
\end{align*}
where $\phi = -\|l_{m,k}-x\|_2^2/(2\sigma^2)$ is a Gaussian kernel. In practice, we set the error deviation $\sigma=0.03$. $Z$ is a normalization constant. 
This contradicts many previous methods on annotating keypoints where a $\delta$-function is implicitly assumed. We argue that it is common that humans make mistakes when annotating keypoints and due to central limit theorem, the keypoint distribution would form a Gaussian.


\subsubsection{Ground-truth Keypoint Generation}
We propose to jointly output a dense mapping function $g_\theta: \mathcal{M} \rightarrow \mathbb{R}^d$ whose parameters are $\theta$, and the aggregated ground-truth keypoint set $\mathcal{Y}$. $g_\theta$ transforms each point into an high-dimensional embedding vector in $\mathbb{R}^d$. Specifically, we solve the following optimization problem:
\begin{align}
\label{eq:main}
\begin{split}
    (\theta^\ast, \mathcal{Y}^\ast) &= \argmin_{\theta, \mathcal{Y}}[f(\mathcal{Y}, g_\theta)  + H(g_\theta)] \\
    &s.t.\ g_\theta(y_{m_1,k}) \equiv g_\theta(y_{m_2,k}), \forall m_1, m_2, k.
\end{split}
\end{align}
where $f(\mathcal{Y}, g_\theta)$ is the data fidelity loss and $H(g_\theta)$ is a regularization term to avoid trivial solution like $g_\theta \equiv 0$, which is defined as $H(g_\theta) = -\sum_{m=1}^M\sum_{k_1,k_2}^{K_m}\|g_\theta(y_{m,k_1})-g_\theta(y_{m,k_2})\|_2^2$. The constraint in Equation~\ref{eq:main} states that the embedding of ground-truth keypoints with the same index should be the same.

\paragraph{Fidelity Loss}
We define $f(\mathcal{Y}, g_\theta)$ as:
\begin{align*}
    f(\mathcal{Y}, g_\theta) = \sum_{m=1}^M\sum_{c=1}^C\sum_{k=1}^{K_m}\int_{\mathcal{M}_m}\mathbf{d}_\theta(x, y_{m,k})\frac{\phi(l_{m,n^*}^{(c)}, x)}{Z(\phi)}dx,
\end{align*}
 where $\mathbf{d}_\theta$ is the L2 distance between two vectors in embedding space:
 \begin{align*}
     \mathbf{d}_\theta(a, b) = \|g_\theta(a) - g_\theta(b)\|_2^2,
 \end{align*}
and
\begin{align*}
    n^* = \argmin_{n} \mathbf{d}_\theta({l_{m,n}^{(c)}}, y_m).
\end{align*}
 
 Unlike previous methods such as Dutagaci et al.\cite{dutagaci2012evaluation} where a simple geodesic average of human labeled points is given as ground-truth points, we seek a point whose expected embedding distance to all human labeled points is smallest. The reason is that geodesic distance is sensitive to the misannotated keypoints and could not distinguish closely spaced keypoints, while embedding distance is more robust to noisy points as the embedding space encodes the semantic information of an object. 
 

\paragraph{Alternating Minimization}
Equation~\ref{eq:main} involves both $\theta$ and $\mathcal{Y}$ and it is impractical to solve this problem in closed form. In practice, we use alternating minimization with a deep neural network to approximate the embedding function $g_\theta$, so that we solve the following dual problem instead (by slightly loosening the constraints):

\begin{align}
\begin{split}
\label{eq:alty}
    \mathcal{Y}^\ast = \argmin_\mathcal{Y}& \sum_{m=1}^M\sum_{c=1}^C\sum_{k=1}^{K_m}\int_{\mathcal{M}_m}\mathbf{d}_\theta(x, y_{m,k})\frac{\phi(l_{m,n^\star}^{(c)}, x)}{Z(\phi)}dx,\\
    &s.t.\ g_\theta(y_{m_1,k}) \equiv g_\theta(y_{m_2,k}), \forall m_1, m_2, k;
\end{split}
\end{align}

\begin{align}
\begin{split}
\label{eq:altth}
    \theta^\ast = \argmin_\theta [H(g_\theta) + \lambda \sum_{m_1,m_2}^M\sum_k^{K_m}\|g_\theta(y_{m_1,k}) - g_\theta(y_{m_2,k})\|_2^2],
\end{split}
\end{align}
and alternate between these two equations until convergence.

Equation~\ref{eq:alty} may be hard to solve at the first place since the ground-truth aligned $K_m$ is unknown beforehand. In order to get aligned keypoint indexes, non-minimum suppression is leveraged first to propose a set of candidate unaligned keypoints. Then we utilize DBSCAN, a clustering algorithm to align these keypoints (e.g. keypoints with same semantics are grouped together). 

Specifically, for each model $\mathcal{M}_m$, we first find an optimal unaligned keypoint set $\mathcal{Y}_m^\ast$ as follows:
\begin{align}
\begin{split}
\label{eq:nms}
    \mathcal{Y}_m^\ast &= \argmin_{y_m\in\mathcal{N}(y_m)}f(y_m, g_\theta)\\
    &= \argmin_{y_m\in\mathcal{N}(y_m)}\sum_{m'\neq m}^M\sum_{c=1}^C\int_{\mathcal{M}_{m'}}\mathbf{d}_\theta(x, y_{m'})\frac{\phi(l_{{m'},k^*}^{(c)}, x)}{Z(\phi)}dx,
\end{split}
\end{align}
where $y_{m'} = \argmin_{y_{m'}\in\mathcal{M}_{m'}}\|g_\theta(y_{m'}) - g_\theta(y_{m})\|_2^2$ and $l_{{m'},k^*}^{(c)}$ is the nearest keypoint of $y_{m'}$ on $\mathcal{M}_{m'}$ labeled by annotator $c$. $\mathcal{N}(y_m)$ is the neighbor points of $y_m$ on model $\mathcal{M}_m$.

Intuitively, Equation~\ref{eq:nms} is conducting a non-minimum suppression (NMS) operation on the fidelity loss $f(y_m, g_\theta)$, where the fidelity loss of $y_m$ is measured by transferring $y_m$ to a corresponding point $y_{m'}$ on all other models via the embedding function $g_\theta$, and calculating $y_{m'}$'s deviation from the human annotation $l_{{m'},k^*}^{(c)}$.



After NMS, we would get several unaligned potential keypoints $y_{m,1}, y_{m,2}, \dots, y_{m,k}$ for each manifold $\mathcal{M}_m$. However, the arbitrarily assigned index $k$ within each model does not provide a consistent semantic correspondence across different models (e.g. $g_\theta(y_{m_1},k)\not\equiv g_\theta(y_{m_2},k)$). Therefore we cluster these points according to their embeddings, by first projecting them onto 2D subspace with t-SNE~\cite{maaten2008visualizing}, as shown in Figure~\ref{fig:airplane} and \ref{fig:car}. In this way, we are able to give an aligned set of keypoints and $K_m$ can now be computed as the number of keypoint clusters.

Upon the convergence of alternating minimization of Equation~\ref{eq:alty} and \ref{eq:altth}, we have an optimal embedding function $g_\theta$, together with intra-class consistent ground-truth keypoints $\mathcal{Y}$, while keeping its embedding distance from human-labeled keypoints as close as possible. The ground-truth keypoints can be viewed as the projection of human labeled data onto embedding space.

\paragraph{Human Verification}
Though the above method automatically aggregate a set of potential set of keypoints with high precision, it omits some keypoints in some cases. As the last step, experts manually verify these keypoints based on some simple priors such as the rotational symmetry and centrosymmetry of an object.

\subsubsection{NMS and Clustering Details}
\begin{figure}[ht]
    \centering
    \includegraphics[width=\columnwidth]{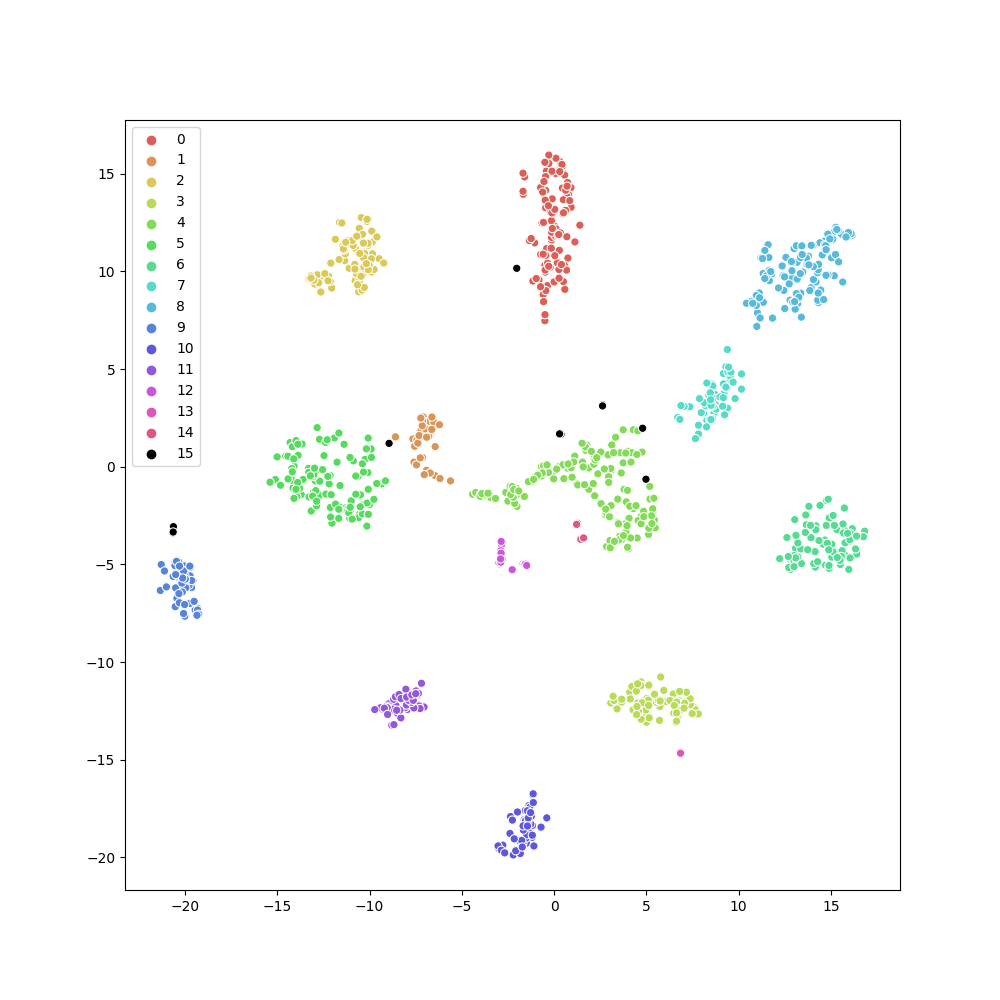}
    \caption{\textbf{Keypoint clustering results for airplanes with DBSCAN.}}
    \label{fig:airplane}
\end{figure}

\begin{figure}[ht]
    \centering
    \includegraphics[width=\columnwidth]{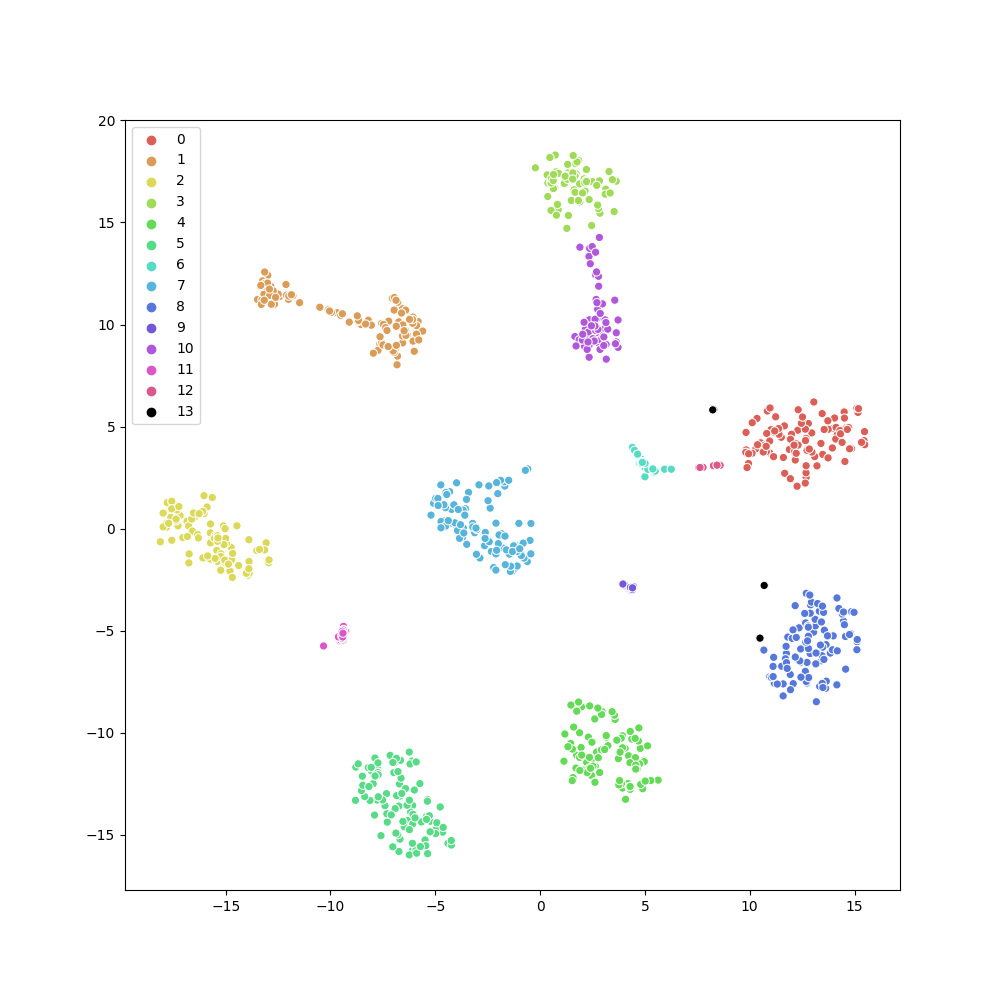}
    \caption{\textbf{Keypoint clustering results for cars with DBSCAN.}}
    \label{fig:car}
\end{figure}
In NMS, we first apply a Gaussian filter to filter our noisy values and then a candidate point $y_m$ is proposed if its fidelity loss $f(y_m, g_\theta)$ is the minimum among those of geodesic neighborhoods. We consider a neighborhood $\mathcal{N}(y_m)$ of size 55 for each point $y_m$. After NMS, we project the embeddings of potential keypoints onto 2D subspace by t-SNE, and then utilize DBSCAN~\cite{ester1996density} algorithm available from \textit{scikit-learn} to do clustering. DBSCAN is short for Density-Based Spatial Clustering of Applications with Noise. It is quite suitable for our keypoint aggregation as it can filter out those noisy keypoints that are accidentally mis-annotated by persons. Clustering results for airplanes and cars are shown in Figure~\ref{fig:airplane} and \ref{fig:car}. Black points indicate noisy points. We can see that potential keypoints are seperated in t-SNE embedding subspace clearly. After clustering, the semantic index of each keypoint could be extracted.

\subsubsection{Optimization Details}
At the start of the alternating minimization, we initialize $\mathcal{Y}$ to be sampled from raw annotations and then run one iteration, which is enough for the convergence. The optimization of Equation~\ref{eq:altth} can be seen as pushing keypoints of different indexes further, while pulling keypoints of the same index closer. Therefore, it can be solved as a classification problem (e.g. logits from the same class get pulled while logits from different classes get pushed). Specifically, we classify point $y_{m,k}$ into $K_m$ classes with a SoftMax layer, and extract the feature of the last but one layer as the embedding $g_\theta$ once the network converged. We choose PointConv as the classification network with dimension 128 in the last but one layer. The learning rate is 1e-3 and the optimizer is Adam~\cite{kingma2014adam}.
\subsubsection{Aggregation Pipeline}
The whole pipeline is shown in Figure~\ref{fig:pipeline}. We first infer dense embeddings from human labeled raw annotations. Then fidelity error maps are calculated by summing embedding distances to human labeled keypoints. Non Minimum Suppression is conducted to form a potential set of keypoints. These keypoints are then projected onto 2D subspace with t-SNE and verified by humans.

\subsection{Symmetries}
In addition to original aggregated keypoints, we manually annotate those points that form a reflection/rotational symmetry group. In particular, as many objects has a reflection plane like chairs and tables as shown in Figure~\ref{fig:vis}, we mark out those reflection point pairs. For square tables in the last row of Figure~\ref{fig:vis}, the points on tabletop form a finite rotational group; while for round tables, we gather the points on tabletop into an infinite rotational group.

\section{Predicting Dense 2D Semantic Embeddings via 3D Knowledge Engine}
\label{sec:pipeline}

\subsection{Transfer 3D semantics to 2D Images by Projection - A Pipeline}
In this section, we describe how to leverage KeypointNet labels with predicted dense embeddings on 3D models, and project them onto 2D images. Hopefully, these embeddings are more consistent since direct inference from 3D models does not suffer from partial visibility and self-occlusion, which often exists on 2D images.

Our method includes four essential parts: (a) 3D shape reconstruction from 2D images, (b) viewpoint estimation from 2D images, (c) 3D semantic embedding prediction from KeypointNet database, (d) 2D projections of 3D semantic embeddings. Full pipeline is illustrated in Figure~\ref{fig:app}.

For 3D shape reconstruction, we utilize a similar structure with ShapeHD~\cite{wu2018learning}, which first estimates silhouettes, normals and depths from 2D images and then predict 3D shapes using 3D convolutions. This pattern can be summarized as 2D-2.5D-3D and is first proposed by MarrNet\cite{wu2017marrnet}. For single-view viewpoint estimation, azimuth and elevation are classified into discrete bins with an ResNet architecture. For 3D semantic embedding prediction, we utilize KeypointNet~\cite{you2020keypointnet} annotations and propose to learn dense embeddings by optimizing contrastive loss among keypoints. For 2D semantic projection, recontructed 3D shapes are projected onto 2D image films with estimated viewpoints, fine-tuned by a differentiable renderer. We follow Pix3D~\cite{wu2018learning} to assume that objects are centered in the image.

\subsection{Single View 3D Reconstruction}
There are a number of works focusing on single view 3D reconstruction, such as MarrNet~\cite{wu2017marrnet}, ShapeHD~\cite{wu2018learning}, Mesh R-CNN~\cite{gkioxari2019mesh}. We utilize a similar architecture with ShapeHD, considering that it could penalize those unrealistic 3D shapes. To make it complete, here we briefly show the components that are used in ShapeHD. ShapeHD is inspired by MarrNet. It has an 2.5D sketch estimator, which is an encoder-decoder that predicts the object's depth, surface normals and silhouette from an RGB image. Followed is a 3D estimator which also has an encoder-decoder structure. The encoder part consists of several 2D convolutional layers to encode previous 2.5D depth, normals and silhouette. The decoder part contains several 3D transpose convolutional layers, which outputs a real value between $[0, 1]$ on a predefined $128^3$ binary voxel grids. Then these grids are converted to 3D meshes through the Marching Cubes algorithm~\cite{lorensen1987marching}. Notice that we predict the 3D shape of the object in the canonical view, as the viewing angle will be estimated by a separate module to be discussed in the next section. Objects in the canonical pose are aligned to the same orientation (e.g. Figure~\ref{fig:vis}) and have unique voxel representations, regardless of how they are positioned in the real scene.

\begin{figure*}[ht]
    \centering
    \includegraphics[width=0.8\linewidth]{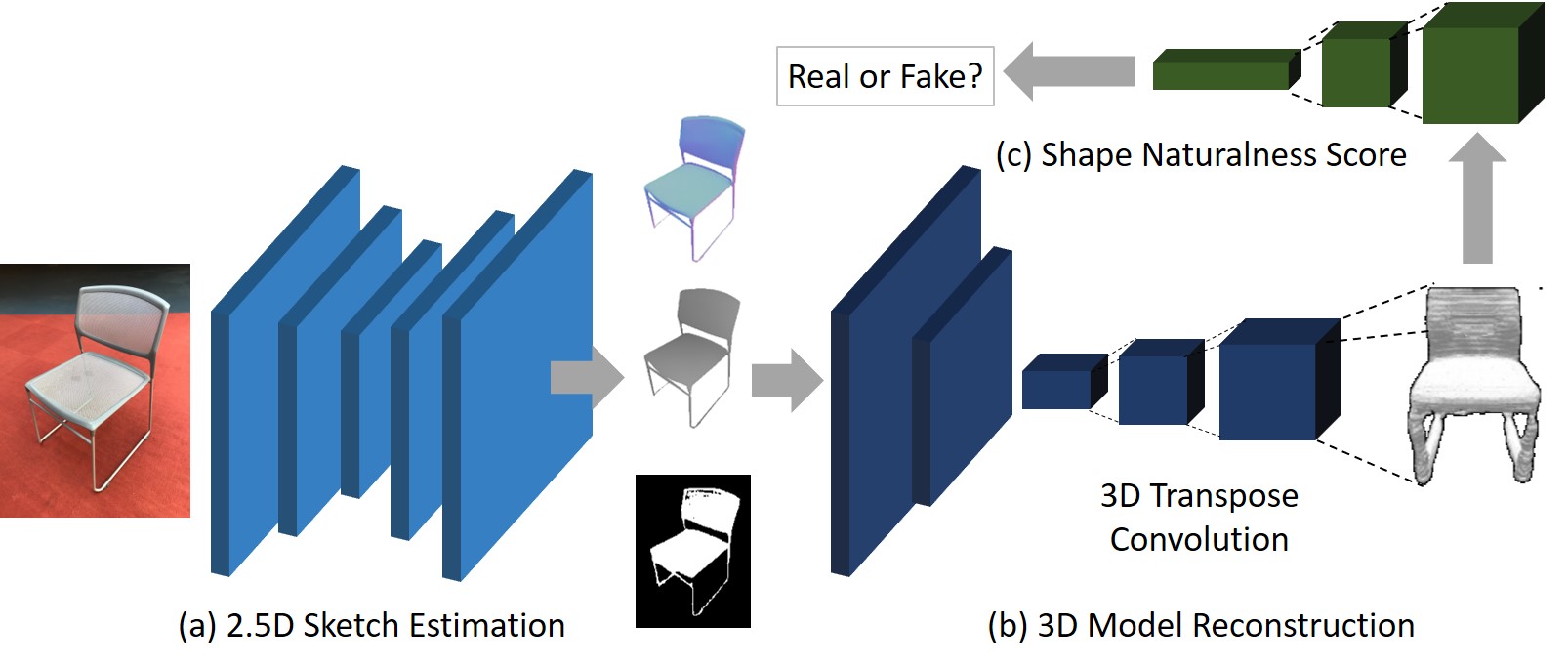}
    \caption{\textbf{Single view 3D model reconstruction}. (a) Firstly, 2.5D sketches including normals, silhouettes and depths are estimated. (b) Then, 3D transpose convolution is used to recover object voxels. (c) In addition, a shape naturalness score is proposed to ensure that generated shapes are not diverged from real shapes.}
    \label{fig:3dreconstruction}
\end{figure*}

In addition, ShapeHD~\cite{wu2018learning} also introduces a deep naturalness regularizer that penalizes unrealistic shapes prediction. The regularizer is implemented by a 3D generative adversarial network and the discriminator is then used to calculate the naturalness score.
The architecture for this module is shown in Figure~\ref{fig:3dreconstruction}.

\subsection{Viewpoint Estimation from RGB Images}
\label{sec:viewpoint}
Given predicted 3D shapes, it is necessary to estimate the viewpoint in order to project it back onto the image plane. To do so, we design a network that predict viewpoints directly from RGB images. Note that this is different from Pix3D~\cite{sun2018pix3d} where viewpoints are estimated from 2.5D sketches. We argue that error would accumulate if the previous 2.5D sketch prediction is inaccurate. Direct estimation reduces the number of passed stage from two to one, which help improve the accuracy. 

\begin{figure}[ht]
    \centering
    \includegraphics[width=\linewidth]{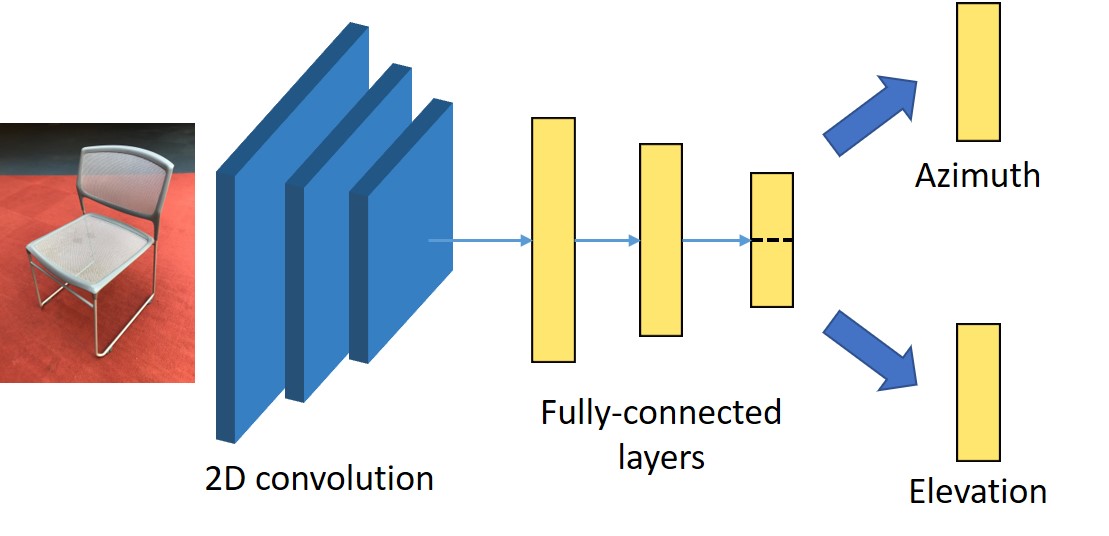}
    \caption{\textbf{Viewpoint Estimation Module}. Viewpoint is estimated from 2D CNN followed by several fully connected layers. Azimuth and elevation are discretized into bins and optimized with cross-entropy loss.}
    \label{fig:viewpoint}
\end{figure}

We treat view estimation as a classification problem, where azimuth is divided into 24 bins the elevation is divided into 12 bins. Circularity in azimuth is dealt carefully with an additional circular bin.
The architecture for this module is demonstrated in Figure~\ref{fig:viewpoint}.

\subsection{3D Semantic Embedding Prediction}
Predicting semantic on 3D models is pretty challenging in this 2D-3D-2D loop. Firstly, the predicted 3D shape from previous stage is not perfect and may be corrupted. Secondly, directly training on 3D models may be prohibitive as current semantic image datasets usually do not come up with the corresponding 3D models. Even for datasets with 3D models like PASCAL-3D~\cite{xiang_wacv14}, since the number of models is relatively small, overfitting is highly suspected. 

Therefore, we resort to previously built KeypointNet dataset. It should be mentioned though only dozens of sparse keypoints per model are annotated in KeypointNet, we could still obtain dense semantic embeddings for each point by leveraging contrastive loss, since human annotated keypoints often provide rich semantics and good coverage over the whole object (e.g. Figure~\ref{fig:vis}). To be specific, keypoints with same semantic should be pulled in the embedding space while keypoints with different semantics are pushed further:

\begin{align}
    L_{contr} = &\sum_{m_1,m_2}^M\sum_{k=1}^K\|h(y_{m_1,k}) - h(y_{m_1,k})\|_2^2 \label{eq:pos}\\
    &- \sum_{m}^M\sum_{k_1, k_2}^K\min(\|h(y_{m,k_1}) - h(y_{m,k_2})\|_2, \delta)^2, \label{eq:neg}
\end{align}

where $K$ is the number of semantic indexes, $M$ is the number of models, $y$ is the aggregated keypoint in Section~\ref{sec:aggr}, $h$ is the embedding network. We would like to learn embeddings with a small distance for positive pairs (Equation~\ref{eq:pos}), and a greater distance than some margin value $\delta$ for negative pairs (Equation~\ref{eq:neg}). In practice, this loss is optimized in an online batch estimation, with hardest negative pair selection. Note that here $h$ has nothing to do with $g_\theta$ as  $g_\theta$ is merely used for the aggregation of unaligned raw keypoint annotations in order to build KeypointNet, while $h$ is trained to generate embeddings on the already aggregated and human verified KeypointNet dataset.
 
\begin{figure*}[ht]
    \centering
    \includegraphics[width=0.7\linewidth]{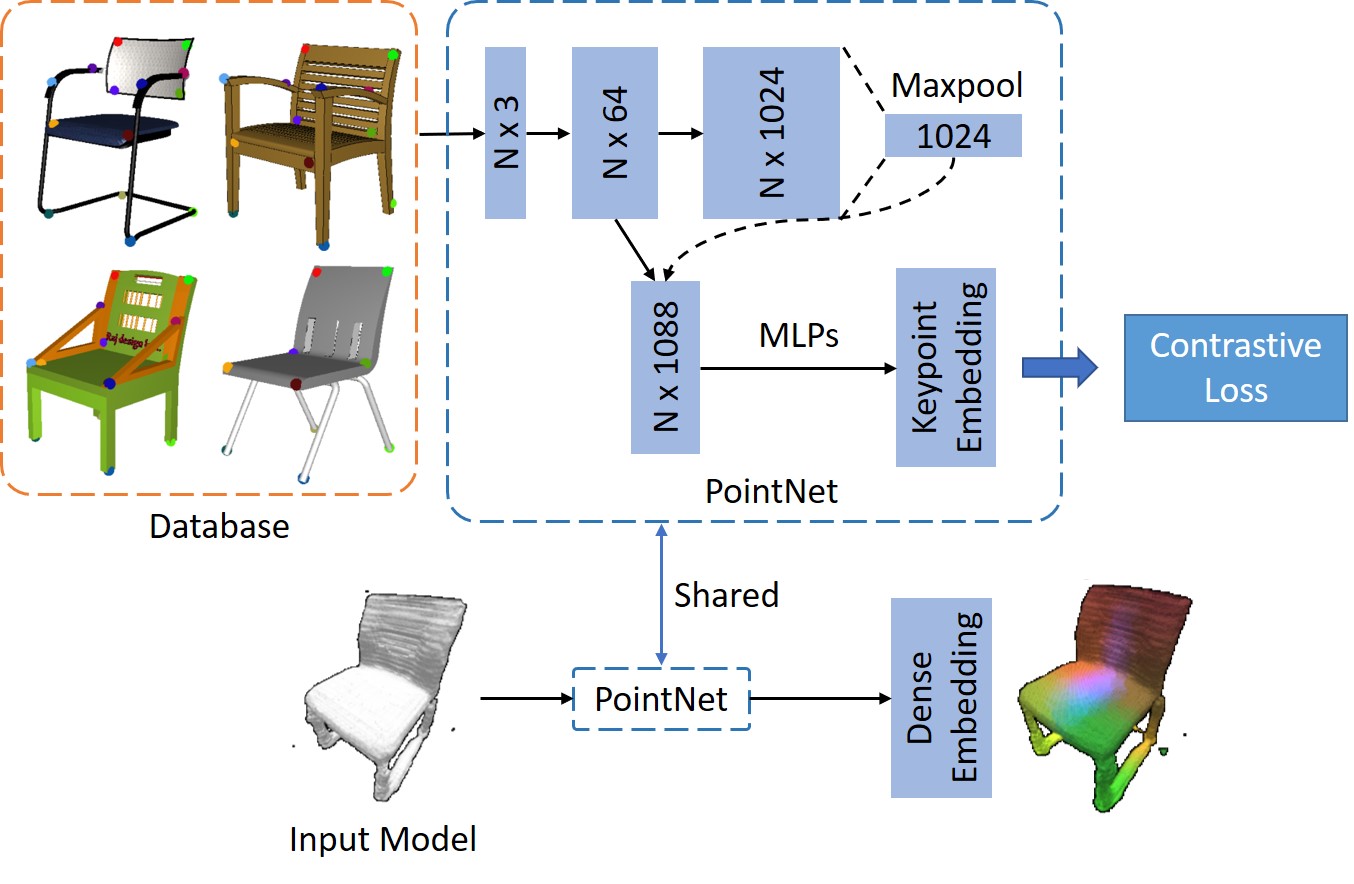}
    \caption{\textbf{3D dense semantic embedding prediction}. The database has a large collection of models (may not necessarily contain the model to be evaluated). We extract their keypoint embeddings using PointNet, whose embeddings are trained with contrastive loss. For an input model, its dense embedding is extracted with the same pretrained PointNet.}
    \label{fig:semantictransfer}
\end{figure*}

We leverage a PointNet-like network structure to extract embeddings on each point. Since our neural network is continuous resulted from the composition of simple continuous layers, it could generalize embeddings on those non-keypoints even though the contrastive loss is defined on keypoints only. In other words, we train a semantic prediction network on a 3D object database and then generalize it to our predicted 3D shapes. To account for potential corruption from previous reconstruction, we augment our dataset with random Gaussian noises near the object surface.
This idea is illustrated in Figure~\ref{fig:semantictransfer}.

\subsection{Differentiable Rendering in 2D Projection}
\label{sec:finetune}
As a final step, we are now ready to project our predicted 3D shapes together with inferred dense semantic embeddings and viewpoints back onto 2D image planes. This step, although the last but not the least, is important as errors are accumulated all the way through previous stages. To have a chance correcting previous predictions, we threshold the voxels with marching cube algorithm, and fine-tune the viewpoint with the help of Neural Mesh Renderer~\cite{kato2018neural}. Specifically, denote the ground-truth silhouette image as $S$, predicted 3D model as $M$, our fine-tuned viewpoint is $V^*$:

\begin{align}
    V^* = \argmin_V\sum_{p\in \Omega}(S[p] - \mathbf{1}(p \in \{K\cdot(R_V\cdot x+T_V)|x\in M\}))^2,
\end{align}

where $\Omega$ is the set of all 2D image coordinates in projective space; $K$ is the fixed camera intrinsic parameters; $R_V$ and $T_V$ are the rotation and translation components of the camera extrinsic parameters, respectively; $\mathbf{1}(\cdot) \in \{0, 1\}$ is the indicator function. We run several gradient descent steps (e.g. 20) in order to find the best viewpoint.

To summarize, we first predict 3D shapes and viewpoints from single view RGB images; then 3D semantic keypoints or any other semantic information is transferred from an existing 3D model database to the predicted 3D shapes; finally, the predicted 3D shapes together with their semantics are projected onto 2D image planes, with viewpoints fine-tuned.

\begin{figure}[ht]
    \centering
    \includegraphics[width=\linewidth]{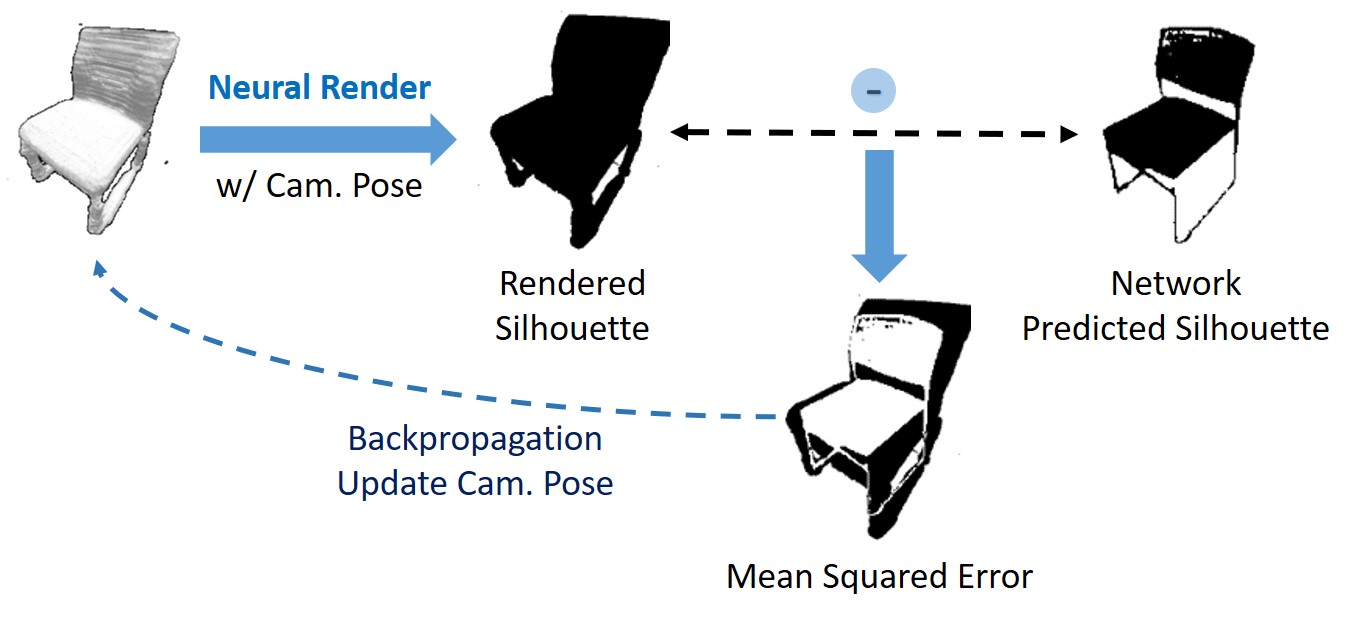}
    \caption{\textbf{Differentiable viewpoint fine-tuning}. Given current camera pose/viewpoints, we back-propagate through neural mesh renderer to optimize its pose by comparing rendered silhouettes and predicted silhouettes.}
    \label{fig:neuralrender}
\end{figure}

\subsection{Experiments}
Our experiments are divided into three parts. The first part is the comparison of our proposed method with current state-of-the-art methods. The second part is ablation studies on our proposed viewpoint estimation and fine-tuning modules.  The third part is a detailed and thorough investigation of each component/stage's influence on final results in our full pipeline. We hope this level of details could help following researchers to have a better understanding on each individual components in the 2D-3D-2D loop.

\subsubsection{Datasets}
\begin{figure}
    \centering
    \includegraphics[width=\linewidth]{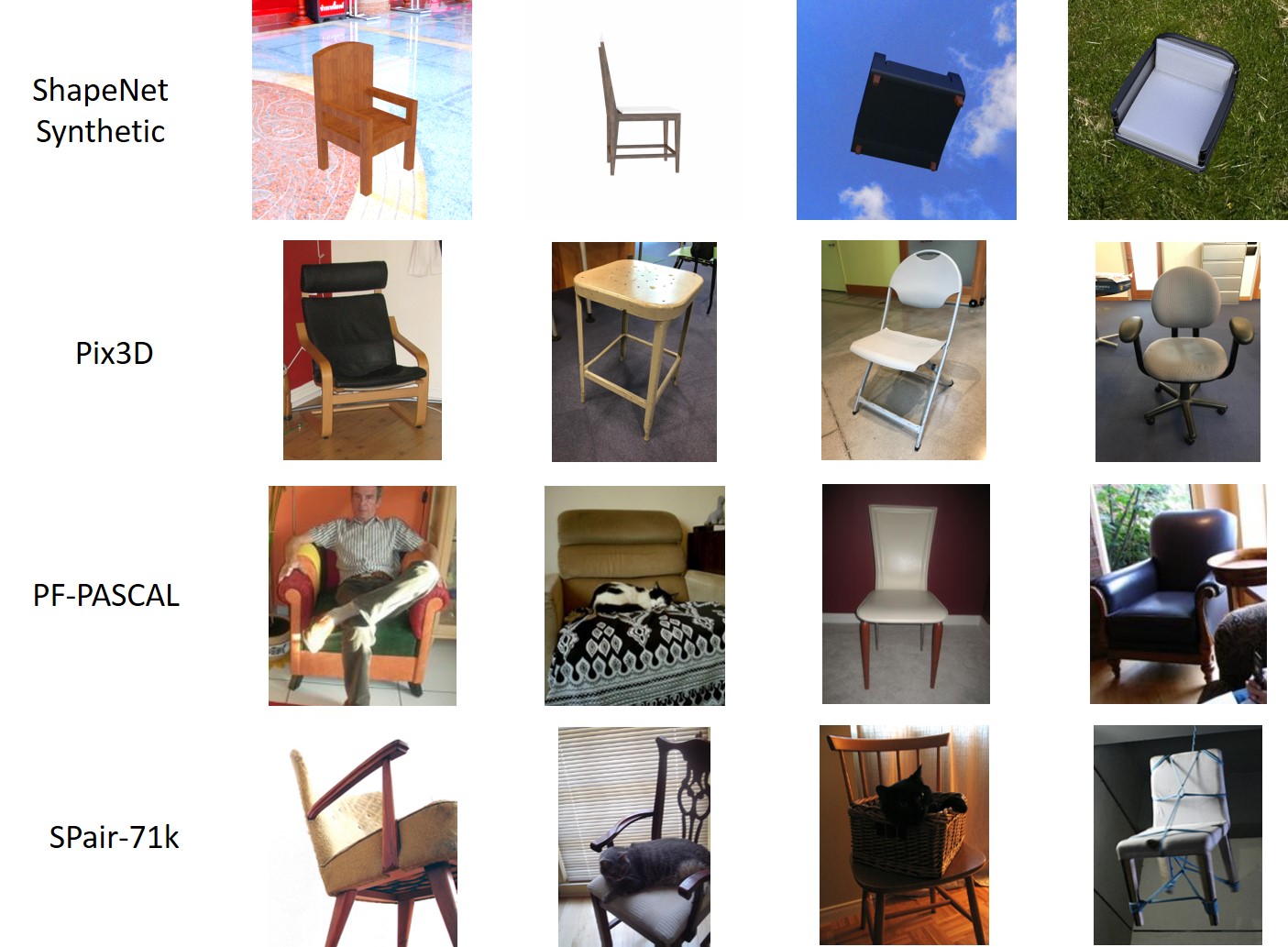}
    \caption{\textbf{Dataset Visualization (chair) on ShapeNet, Pix3D, PF-PASCAL and SPair-71k}. From up to bottom: difficulties from easy to hard. }
    \label{fig:discrep}
\end{figure}

Our method uses ShapeNet Synthetic rendered images~\cite{wu2018learning} for training and Pix3D~\cite{sun2018pix3d}, ~PF-PASCAL\cite{ham2016proposal}, SPair-71k~\cite{min2019spair} images for evaluation. Differences among these four datasets are shown in Figure~\ref{fig:discrep}. ShapeNet Synthetic could provide tons of training data though it is fake and synthetically rendered. Pix3D is a real dataset without much clutter occlusions and objects are well centered in images, which is relatively clean. PF-PASCAL and SPair-71k provide more extreme occlusions/cutoff/scale variations. Though clutter occlusions and incompletions are not the focus of this paper, our method still gives a competitive score on these datasets compared with state-of-the-art methods.

Our 2D-3D single image reconstruction is based on ShapeHD and trained on the ShapeNet Synthetic rendered images released officially by Wu \etal~\cite{wu2018learning}, which contains three common categories: chairs, cars and aeroplanes. Therefore, experimental results on these categories are reported for PF-PASCAL and SPair-71k. For Pix3D, only chairs are evaluated since it does not contain car and aeroplane labels.

\subsubsection{Metric}
We follow previous works to use a common evaluation metric on 2D semantic understanding: percentage of correct keypoints (PCK), which counts the average number of correctly predicted keypoints given a tolerance threshold. Given predicted keypoint $\mathbf{k}_{pr}$ and groundtruth keypoint $\mathbf{k}_{gt}$, the prediction is considered correct if
Euclidean distance between them is smaller than a given
threshold. The correctness $c$ of each keypoint can be expressed as

\begin{align}
    c = \begin{cases}
1 &\text{if\quad$d(\mathbf{k}_{pr},\mathbf{k}_{gt})\le \alpha_\tau\cdot\mathrm{max}(w_\tau,h_\tau)$}\\
0 &\text{otherwise,}
\end{cases}
\end{align}
where $w_\tau$ and $h_\tau$ are the width and height of either an entire image or object bounding box, $\tau\in$\{img, bbox\}, and $\alpha_\tau$ is a tolerance factor.

For a fair comparison, we select those keypoints that appear in both KeypointNet and Pix3D/PF-PASCAL/SPair-71k for evaluation. For 2D warping methods like HPF~\cite{min2019hyperpixel}, PCK is calculated on each image pair where keypoints in one reference image are reckoned as ground-truth and keypoints in the other image are predicted by a semantic warp or transfer. For 3D transfer methods like ours, keypoints in images are found by projecting keypoints on predicted 3D shapes and PCK is calculated for each individual image. Our method can be also considered as transferring keypoints from an implicit ground-truth 3D keypoint template. PCK results are averaged over all keypoints. 

\subsubsection{Implementation Details}
All our networks are written in Pytorch. Each stage in Figure~\ref{fig:pipeline} is trained independently. Our 2.5D sketch estimation network and 3D completion network are trained
with images rendered with ShapeNet Synthetic objects. We train the 2.5D sketch estimator using a mean squared error loss and SGD with a learning rate of 0.001 for 120 epochs. 3D reconsutruction network is trained with SGD of learning rate 0.1 and momentum 0.9 for 80 epochs. The naturalness network is trained with ADAM of learning rate 0.001 for 80 epochs. Viewpoint estimation network is trained with ADAM of learning rate 0.001 for 100 epochs. All input images are cropped and resized to $480\times 480$ so that the object is centered in the image.


\begin{figure*}[ht]
    \centering
    \includegraphics[width=0.95\linewidth]{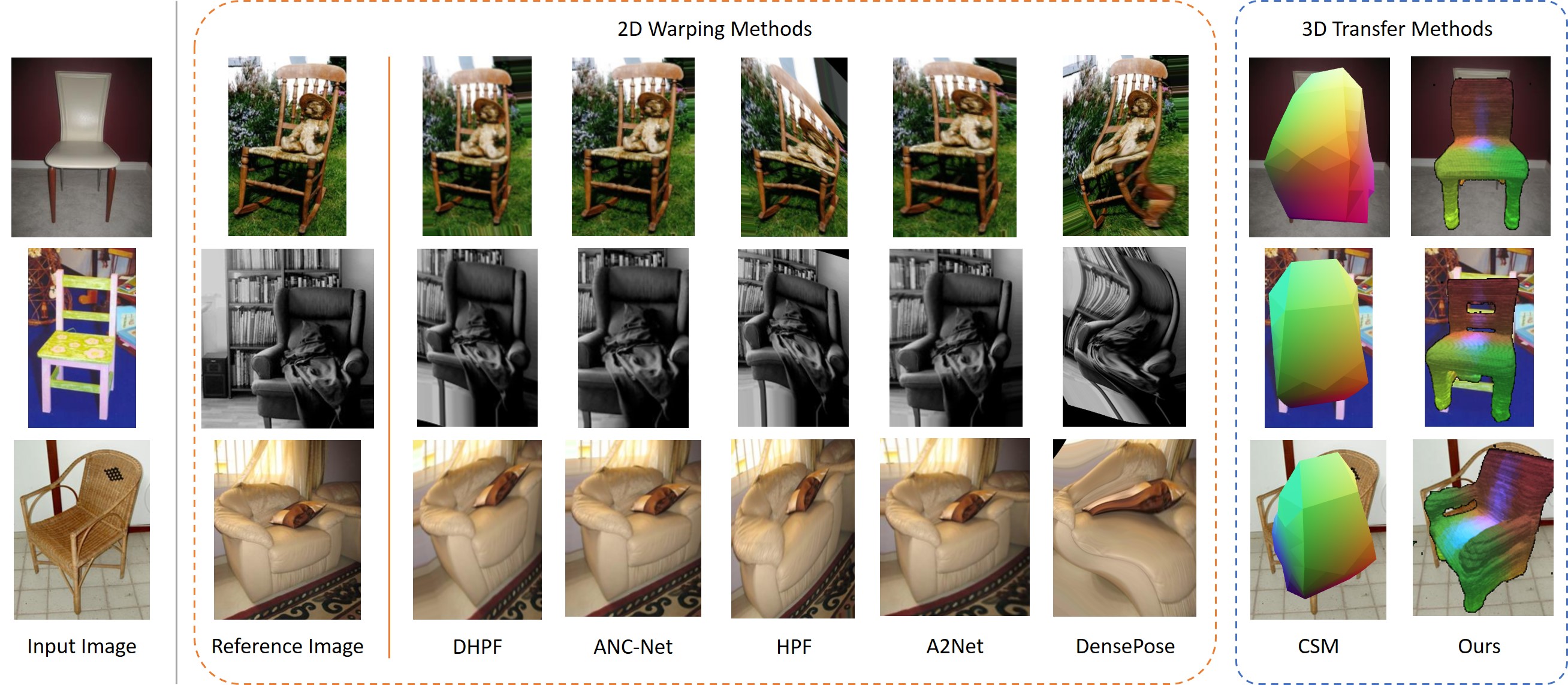}
    \caption{\textbf{Qualitative comparison on chairs in PF-PASCAL}. 2D warping methods fail to align dense semantics from a reference image due to viewpoint variations. CSM assumes a fixed template and fail to project accurate embeddings onto 2D images.}
    \label{fig:pfpascal}
\end{figure*}

\begin{figure*}[ht]
    \centering
    \includegraphics[width=0.95\linewidth]{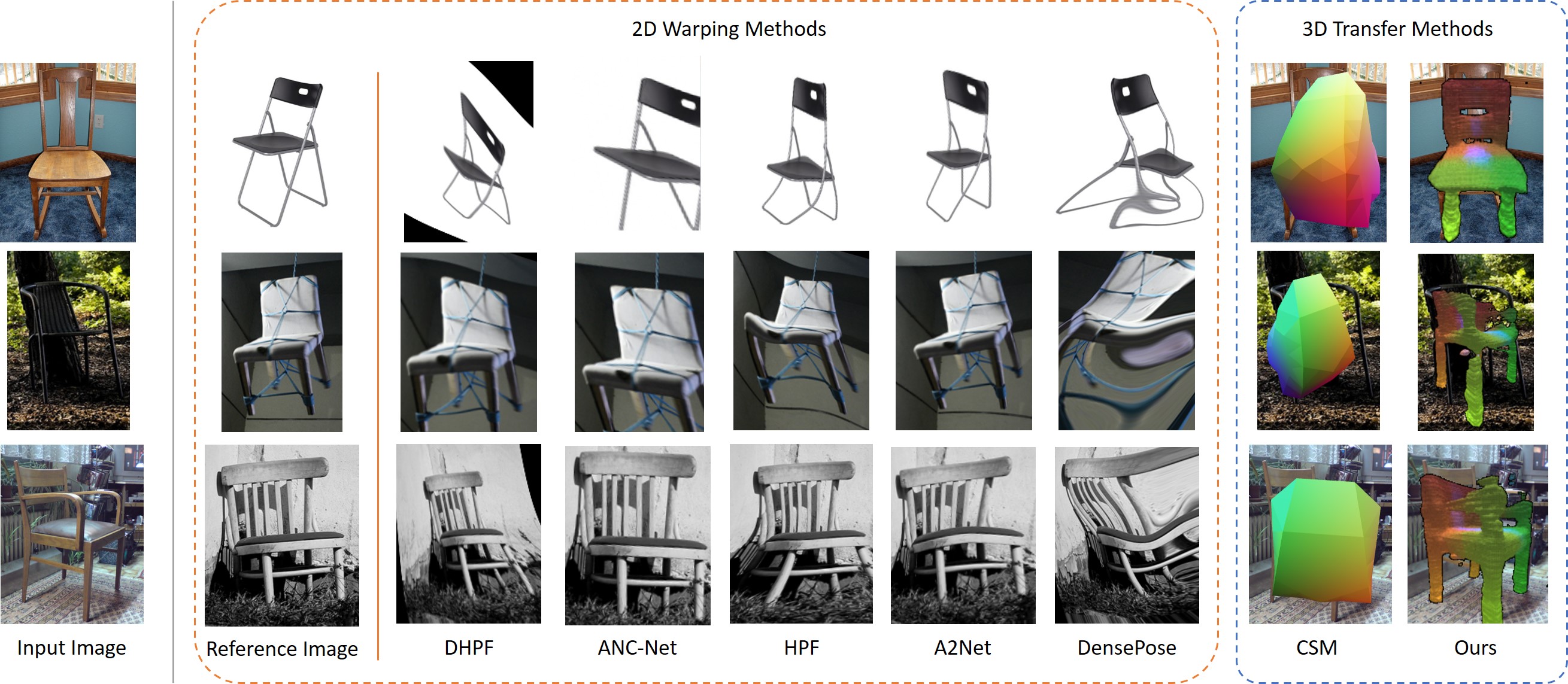}
    \caption{\textbf{Qualitative comparison on chairs in SPair-71k}.}
    \label{fig:spair}
\end{figure*}

\begin{figure*}[ht]
    \centering
    \includegraphics[width=0.8\linewidth]{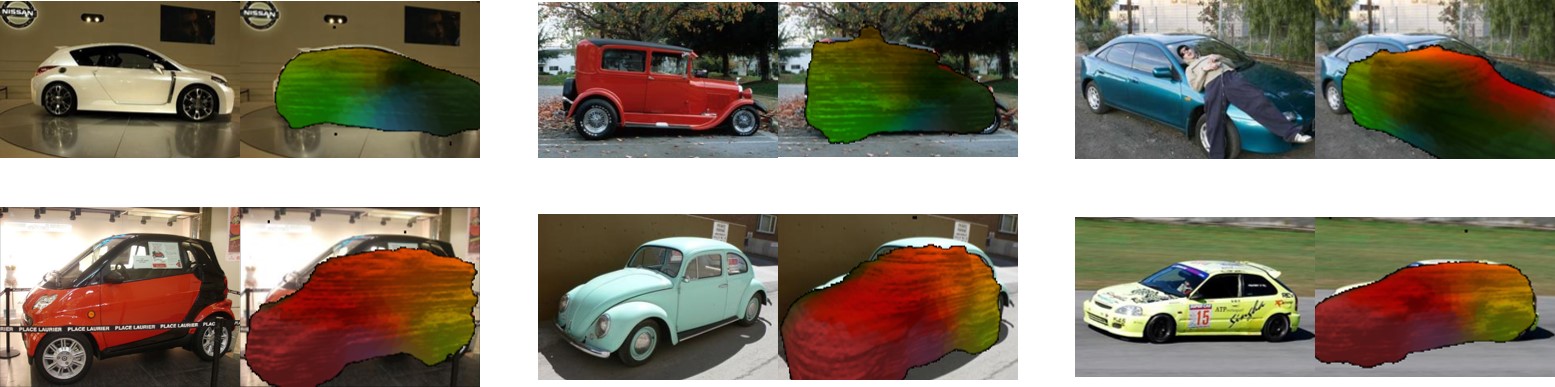}
    \bigbreak
    \includegraphics[width=0.8\linewidth]{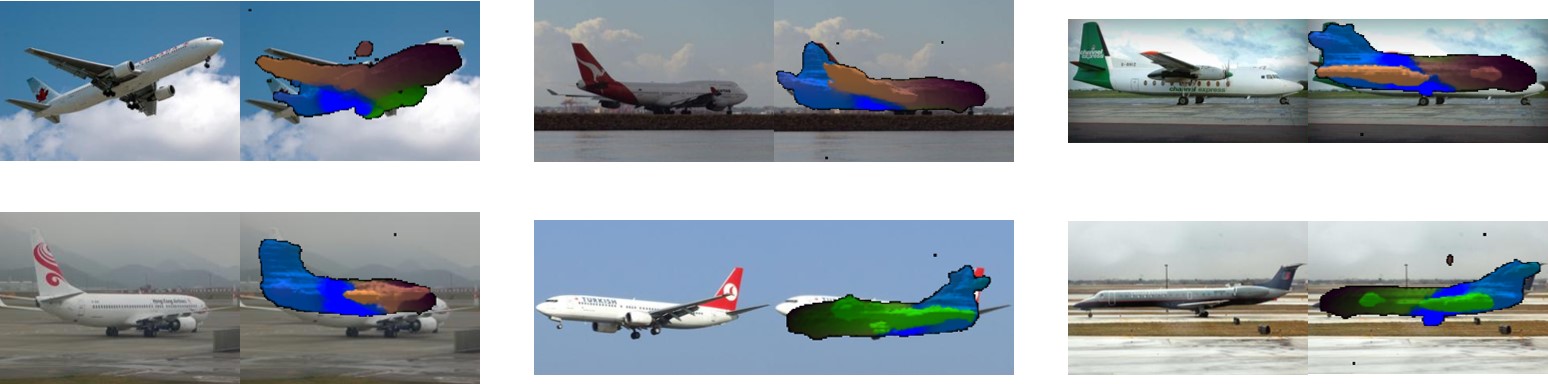}
    \caption{\textbf{Qualitative results of transferred dense embeddings on cars and aeroplanes.} For each image pair, left: original image, right: projected 3D models textured with transferred dense embeddings. Embeddings are visualized as RGB color tuples by PCA. Notice that our model predicts consistent embeddings across different objects under various viewpoints.}
    \label{fig:airplane_emb}
\end{figure*}

    

\begin{table*}[ht]
\centering
\resizebox{0.9\linewidth}{!}{
\begin{tabular}{l|c|c|c|c|c|c|c}
    \hline
    \multirow{2}{*}{Models} & \multicolumn{3}{c|}{PF-PASCAL} & \multicolumn{3}{c|}{SPair71k} & \multicolumn{1}{c}{Pix3D} \\
        \cline{2-8}
        & Chair & Car & Aeroplane & Chair & Car & Aeroplane & Chair\\
    \hline
    \textbf{Ours} & 0.444/0.342 & \textbf{0.574}/0.349 & \textbf{0.440}/0.194 & \textbf{0.565}/\textbf{0.346} & \textbf{0.500}/\textbf{0.328} & 0.390/0.182 & \textbf{0.560}/0.323\\
    \hline
    DHPF\textsubscript{res101}\cite{min2019dynamic} & 0.601/\textbf{0.444} & 0.456/0.356 & 0.384/0.299 & 0.363/0.268 & 0.312/0.227 & \textbf{0.442}/\textbf{0.362} & 0.514/0.424 \\ 
    ANC-Net\textsubscript{res101}\cite{li2020correspondence} & 0.455/0.288 & 0.410/0.302 & 0.391/0.299 & 0.307/0.194  & 0.253/0.183 & 0.264/0.201 & 0.491/0.372 \\ 
    HPF\textsubscript{res101}\cite{min2019hyperpixel} &  \textbf{0.602}/0.435  & 0.533/\textbf{0.437} & 0.401/0.294 & 0.419/0.324 & 0.364/0.276 & 0.280/0.212 & 0.534/\textbf{0.434}\\
    A2Net\textsubscript{res101}\cite{seo2018attentive} & 0.516/0.302 & 0.478/0.326 & 0.388/0.256 & 0.358/0.200 & 0.332/0.201 & 0.234/0.155 & 0.506/0.356\\
    CSM\textsubscript{unet}\footnotemark[1]\cite{kulkarni2019canonical} & 0.164/0.164 & 0.339/0.339 & 0.220/0.220 & 0.115/0.115 & 0.234/0.234 & 0.139/0.139 & 0.152/0.152\\
    DensePose\textsubscript{unet}\cite{guler2018densepose} & 0.207/0.148 & 0.220/0.158 & 0.397/\textbf{0.301} &   0.212/0.172 & 0.159/0.114 & 0.235/0.190 & 0.323/0.263\\
    \hline
\end{tabular}}
\bigbreak
\caption{\textbf{Comparison of our method with state-of-the-arts}. Each PCK score is reported under both $\alpha_{img}=0.1$ and $\alpha_{bbox}=0.1$, separated by slashes. \textsuperscript{1}CSM crops input images with object bounding boxes, so the results for $\alpha_{img}=0.1$ and $\alpha_{bbox}=0.1$ are the same.}
\label{tab:compsoa}
\end{table*}

\begin{table*}[htbp!]
\centering
\resizebox{0.7\linewidth}{!}{
\begin{tabular}{l|c|c|c}
    \hline
    \multirow{2}{*}{Models} & \multicolumn{1}{c|}{PF-PASCAL} & \multicolumn{1}{c|}{SPair71k} & \multicolumn{1}{c}{Pix3D}  \\
        \cline{2-4}
        & Chair & Chair & Chair \\
    \hline
    Ours & 0.444/\textbf{0.342} & \textbf{0.565}/\textbf{0.346} & \textbf{0.560}/\textbf{0.323}\\
    \hline
    Ours w/ viewpoint from 2.5D sketches  & 0.171/0.075 & 0.185/0.098 & 0.238/0.117\\
    Ours w/o viewpoint fine-tuning & \textbf{0.497}/0.316 & 0.538/0.332 & 0.560/0.322\\
    \hline
\end{tabular}}
\bigbreak
\caption{\textbf{Ablation study}. PCK is reported under both $\alpha_{img}=0.1$ and $\alpha_{bbox}=0.1$, separated by slashes.}
\label{tab:ablation}
\end{table*}

\begin{table}[htbp!]
\centering
\resizebox{0.9\linewidth}{!}{
\begin{tabular}{l|c|c}
    \hline
    \multirow{2}{*}{Models} & \multicolumn{1}{c|}{Pix3D} & \multicolumn{1}{c}{ShapeNet} \\
        \cline{2-3}
         & Chair & Chair\\
    \hline
    Ours &  0.560/0.323 & 0.513/0.243\\
    \hline
    Ours w/ g.t. 2.5D sketches  & -/- & 0.518 /0.243\\
    Ours w/ g.t. 3D models  & 0.571/\textbf{0.363} & 0.521/\textbf{0.263}\\
    Ours w/ g.t. viewpoints  & \textbf{0.606}/0.351 & \textbf{0.545}/0.262 \\
    \hline
\end{tabular}}
\bigbreak
\caption{\textbf{Detailed analysis of each component with PCK results evaluated on Pix3D and ShapeNet chairs}. PCK is reported under both $\alpha_{img}=0.1$ and $\alpha_{bbox}=0.1$, separated by slashes. Notice that Pix3D does not provide ground-truth 2D depths and silhouettes.}
\label{tab:gtpix3d}
\end{table}

\subsubsection{Comparison with State-of-the-Arts}
\label{sec:soa}
In this section, we compare our methods with several state-of-the-arts that either directly do a 2D image semantic transfer like DHPF~\cite{min2019dynamic}, ANC-Net~\cite{li2020correspondence}, HPF~\cite{min2019hyperpixel}, A2Net~\cite{seo2018attentive} and DensePose~\cite{guler2018densepose}, or utilize 3D templates like CSM~\cite{kulkarni2019canonical}.

Our method is trained with ShapeNet Synthetic renderings while state-of-the-art methods are trained directly on real-world images. Interestingly, though our method has a domain gap when applied to real-world images, we still gives competitive performance with state-of-the-art methods. 
Quantitative results are listed in Table~\ref{tab:compsoa}. Qualitative comparison on chairs with previous methods are demonstrated in Figure~\ref{fig:pfpascal} and \ref{fig:spair}. We see that previous 2D warping methods fail to align objects in two images, especially when they are from distinct viewpoints. For CSM, since it assumes a fixed template for all instances in a category, the 3D fixed mean template fails to transfer 3D semantic onto 2D images. Our method takes both viewpoints and 3D instance shapes into consideration and performs better than previous methods. More dense embedding visualization on cars and aeroplanes are given in Figure~\ref{fig:airplane_emb}.


\subsubsection{Ablation Study}
In this section, we explore the effect of proposed viewpoint estimation module (Section~\ref{sec:viewpoint}) and viewpoint fine-tuning module (Section~\ref{sec:finetune}). 

For the view estimation module, we compare with a baseline where viewpoints are predicted from predicted 2.5D sketch instead of 2D input images. Quantitative results are listed in Table~\ref{tab:ablation}, we see that our proposed method greatly improve the accuracy of view estimation by avoiding the error in the 2D-2.5D sketch prediction.

For the viewpoint fine-tuning module, we compare with a baseline without fine-tuning. Viewpoint fine-tuning gives our model an additional chance to correct any imperfection of previous predicted 3D shapes or viewpoints. Quantitative results are also listed in Table~\ref{tab:ablation}.

\subsubsection{Detailed Analysis}
In this section, we investigate how each component influences the final result, by replacing the following three components with their ground-truths: 2D to 2.5D sketch estimation, 2.5D to 3D model reconstruction and viewpoint estimation.

We start with our full pipeline and then replace each prediction module with its ground-truth counterpart to see the accuracy improvement, respectively. Notice that although ground-truth viewpoints with azimuth and elevation are given, they are not ground-truth 6D camera poses due to the misalignment of camera centers.

Results are given in Table~\ref{tab:gtpix3d}. The contribution of ground-truth 3D reconstruction is small, which suggests that the 2D-2.5D-3D single view reconstruction pipeline trained on synthetic ShapeNet dataset meets few difficulties when transferred to real datasets. In contrast, replacing predicted viewpoints with ground-truths gives 8.2\% PCK improvement, which suggests that there is still some potential future improvements in single view camera pose estimation.


\section{Conclusion}
In this paper, we propose a novel method on boosting general object semantic understanding by building a 3D large-scale and high-quality KeypointNet dataset. We predict semantic correspondences by leveraging it to 3D domain and then project corresponding 3D models back to 2D domain, with their semantic embeddings. This method explicitly reasons about objects self-occlusion and visibility. We show that our method gives comparative and even superior results on standard semantic benchmarks.

\ifCLASSOPTIONcompsoc
  \section*{Acknowledgments}
\else
  \section*{Acknowledgment}
\fi
This work is supported in part by the National Key R\&D Program of China, No. 2017YFA0700800, National Natural Science Foundation of China under Grants 61772332, Grant 51675342 and Grant 51975350. This work is also supported by SHEITC (2018-RGZN-02046) and Shanghai Qi Zhi Institute.



\bibliographystyle{IEEEtran}
\bibliography{main}



%

%
\begin{IEEEbiography}
    [{\includegraphics[width=1in,height=1.25in,clip,keepaspectratio]{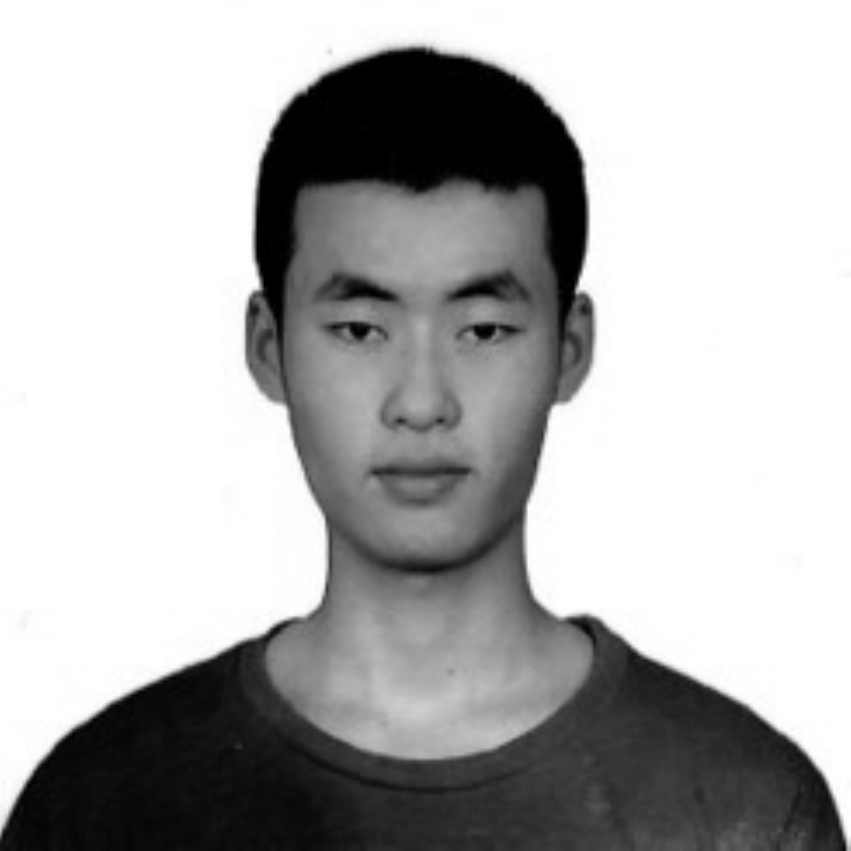}}]{Yang You}
received the BS and MS degrees from Shanghai Jiao Tong University and University of Virginia in 2016 and 2017 respectively. He is now a third-year student pursuing his doctoral degree at Mechanical Engineering department in Shanghai Jiao Tong University. His main interests lie in 3D Computer Vision,
Computer Graphics and Robotics.
\end{IEEEbiography}

\begin{IEEEbiography}
    [{\includegraphics[width=1in,height=1.25in,clip,keepaspectratio]{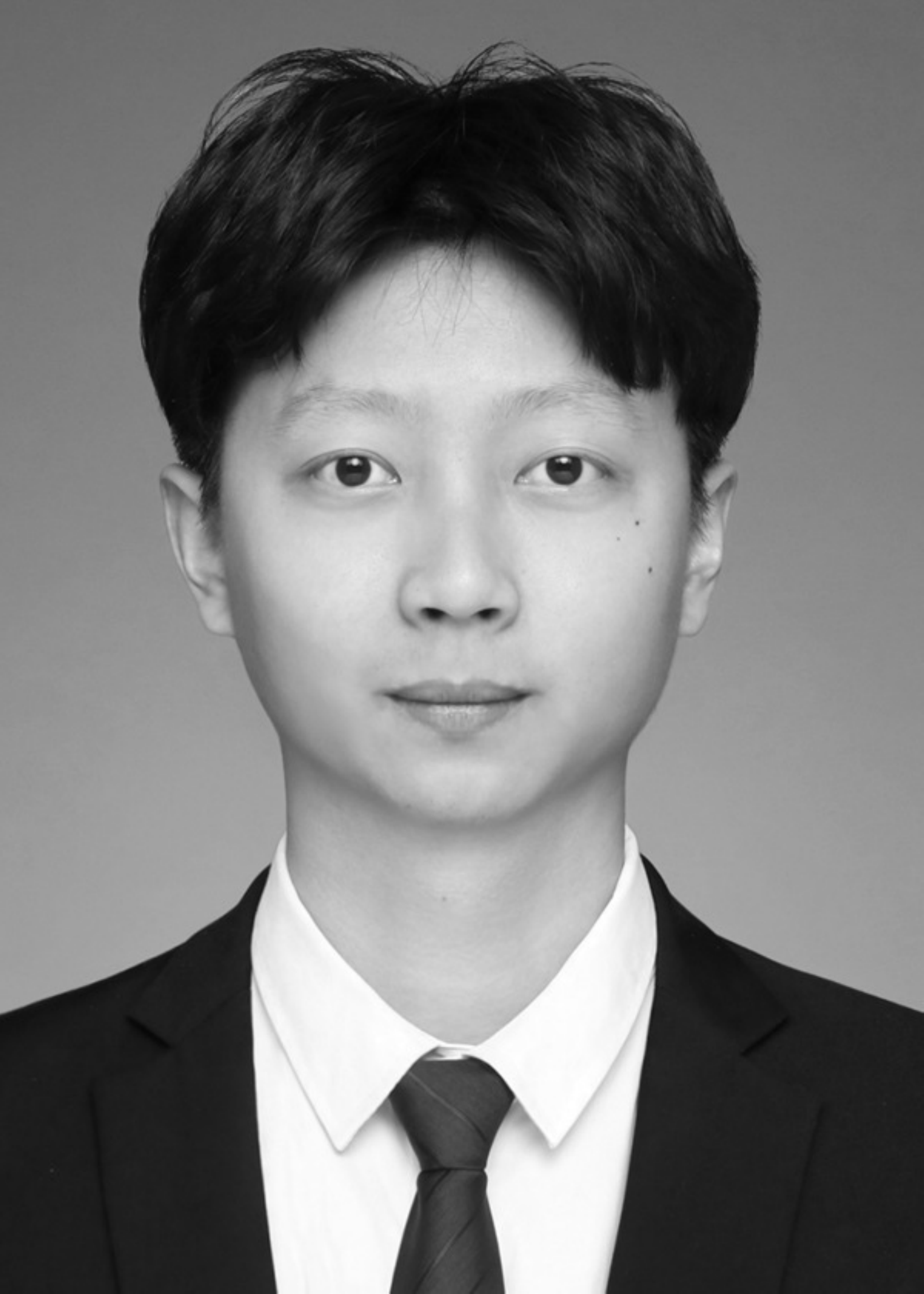}}]{Chengkun Li}
received the BS degree from Shanghai Jiao Tong University in 2018. He is currently pursuing a double master’s degree at Shanghai Jiao Tong University and Ecole Centrale Paris (CentraleSupélec). His current research interests are 3D computer vision, deep learning and robotics.
\end{IEEEbiography}

\begin{IEEEbiography}
    [{\includegraphics[width=1in,height=1.25in,clip,keepaspectratio]{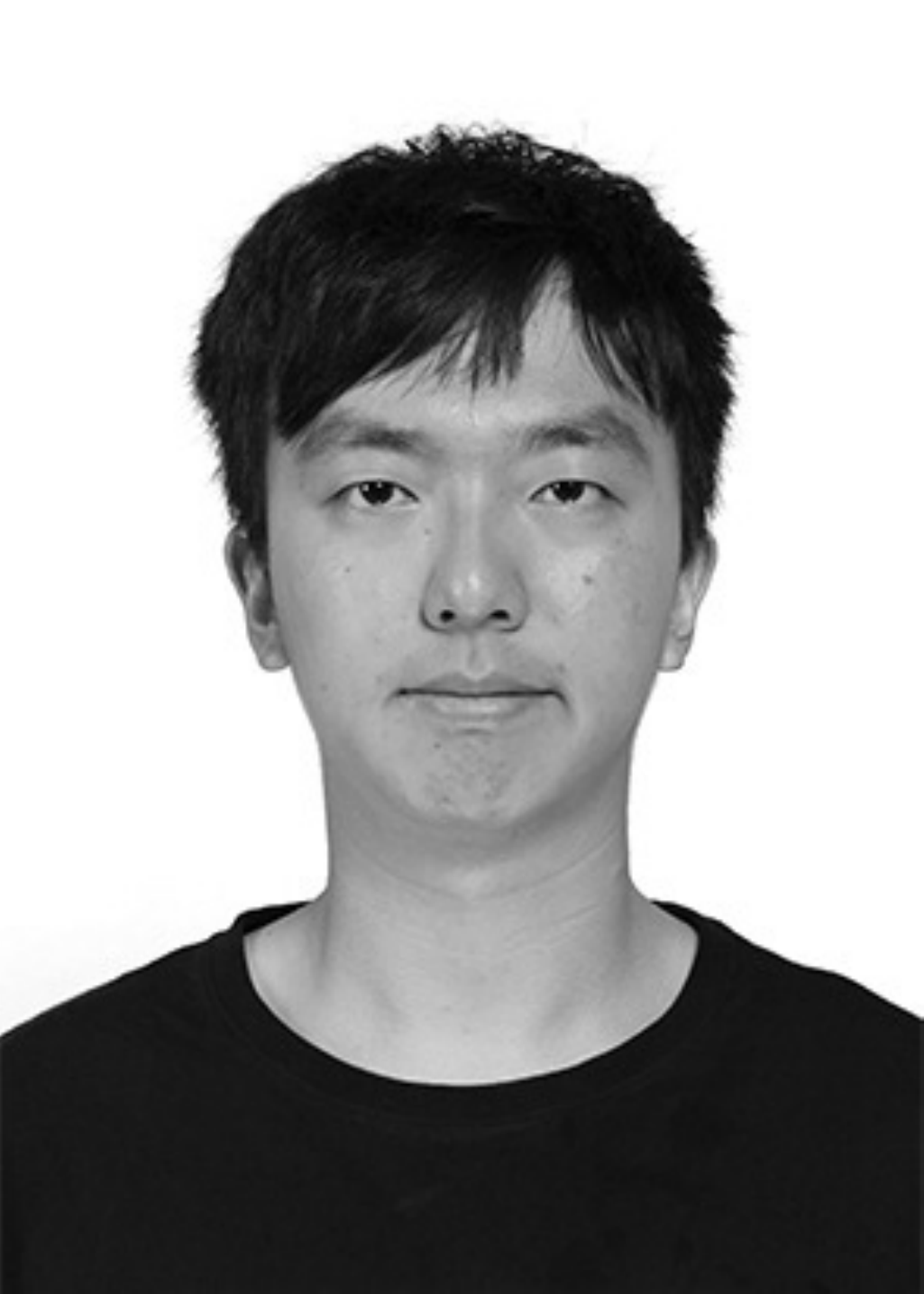}}]{Yujing Lou}
received the B.S. degree in computer science and technology from Harbin Institute of Technology, China, in 2018 and the M.S. degree in computer science and technology from Shanghai Jiao Tong University, China, in 2020. He is currently a Ph.D. student with MVIG lab, Shanghai Jiao Tong University. His research interests include 3D scene/object perception and robot vision.
\end{IEEEbiography}

\begin{IEEEbiography}
    [{\includegraphics[width=1in,height=1.25in,clip,keepaspectratio]{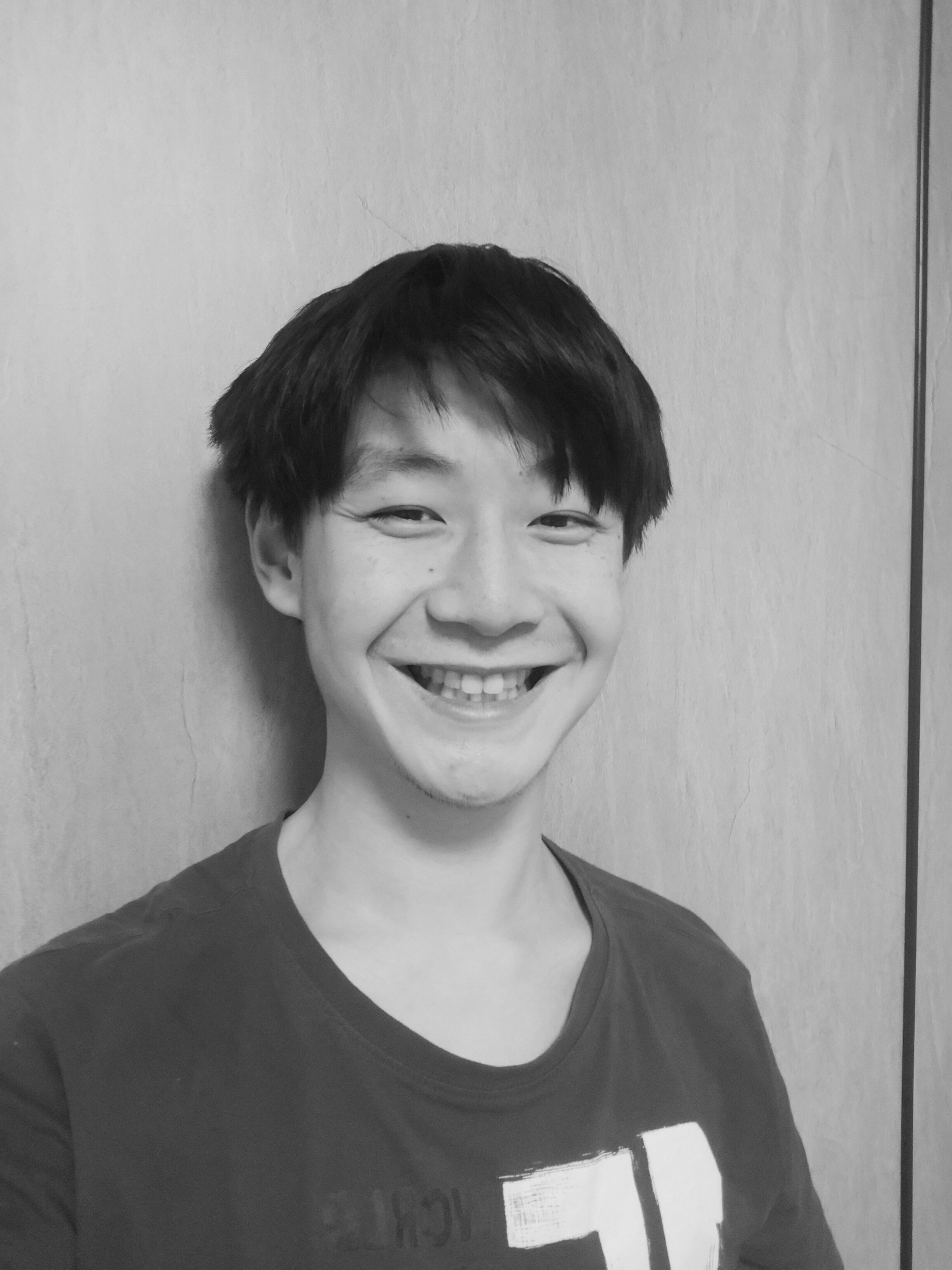}}]{Zhoujun Cheng}
is a senior student  in Shanghai Jiao Tong University.  
He was a intern at Machine Vision and Interlligence Group focusing on 3D shape correspondence under the  supervision of Prof. Cewu Lu in 2019. 
Currently he is working on Table Analysis in Microsoft Research Asia, and will continue his post-graduate study in Information and Computing Lab under the supervision of Prof. Fan Cheng.
\end{IEEEbiography}

\begin{IEEEbiography}
    [{\includegraphics[width=1in,height=1.25in,clip,keepaspectratio]{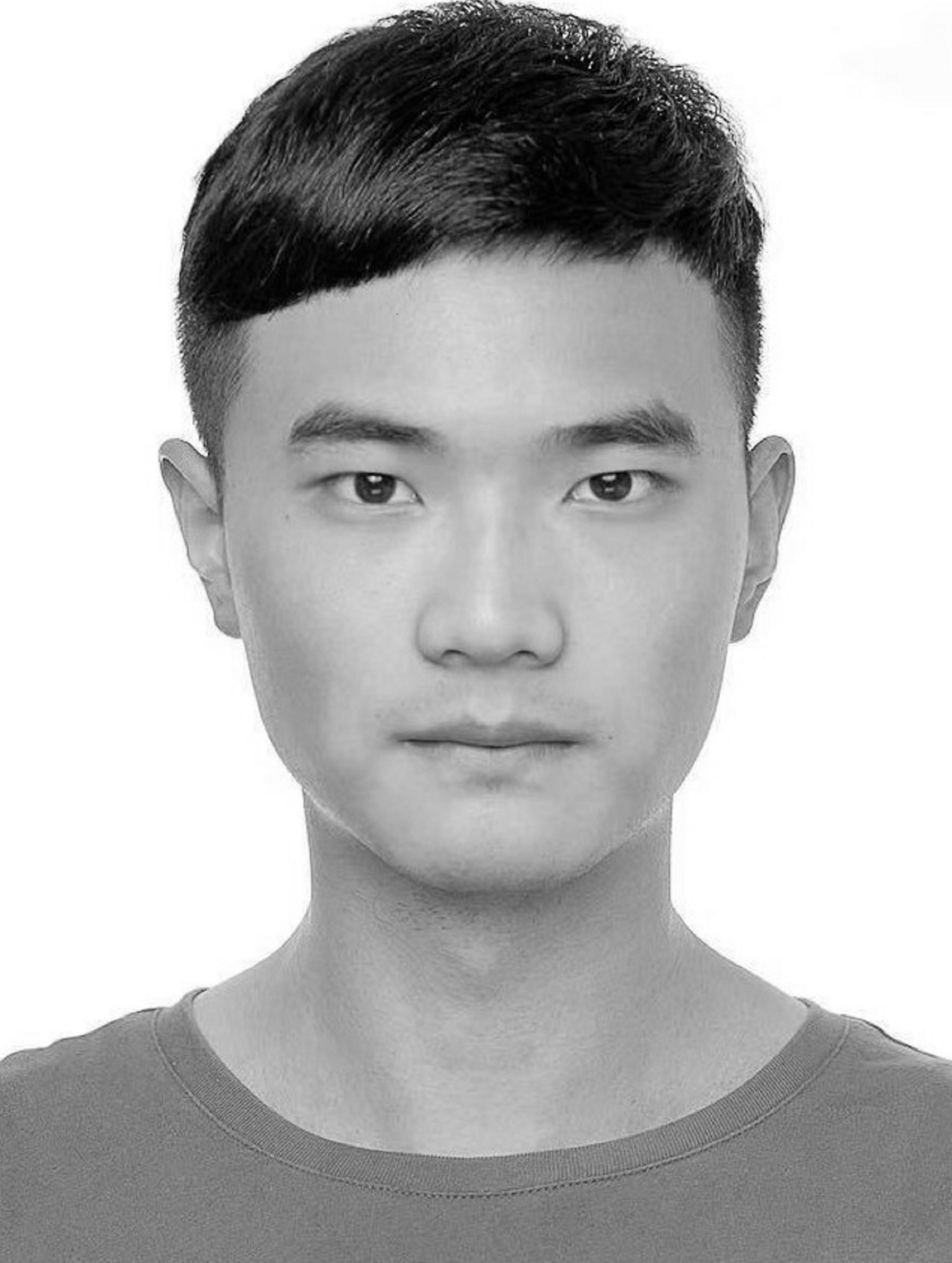}}]{Liangwei Li}
received the BS degree from Shanghai Jiao Tong University (SJTU), China, in 2020. He was an undergraduate researcher in Machine Vision and Intelligence Group (MVIG), Department of Computer Science and Engineering, SJTU. He currently works as an algorithm engineer at Alibaba-inc, China. His research interests include data mining, knowledge reasoning and deep learning applied to recommendation system. He has authored 4 research papers. 
\end{IEEEbiography}

\begin{IEEEbiography}
    [{\includegraphics[width=1in,height=1.25in,clip,keepaspectratio]{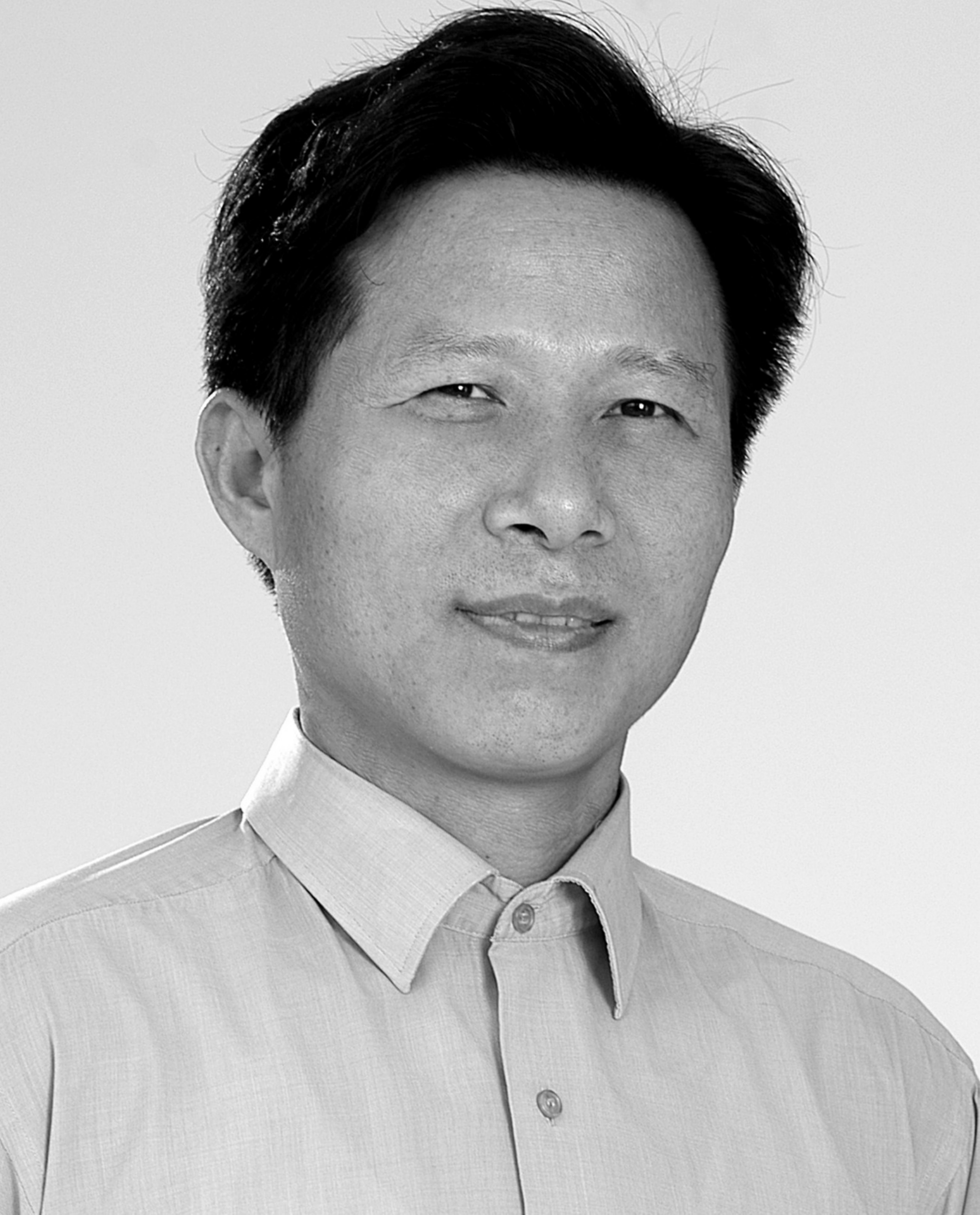}}]{Lizhuang Ma}
received his B.S. and Ph.D. degrees from the Zhejiang University, China in 1985 and 1991, respectively. He is now a Distinguished Professor, at the Department of Computer Science and Engineering, Shanghai Jiao Tong University, China and the School of Computer Science and Technology, East China Normal University, China. He was a Visiting Professor at the Frounhofer IGD, Darmstadt, Germany in 1998, and a Visiting Professor at the Center for Advanced Media Technology, Nanyang Technological University, Singapore from 1999 to 2000. 
His research interests include computer vision, computer aided geometric design, computer graphics, scientific data visualization,
computer animation, digital media technology, and theory and applications for computer graphics, CAD/CAM.
\end{IEEEbiography}

\begin{IEEEbiography}
    [{\includegraphics[width=1in,height=1.25in,clip,keepaspectratio]{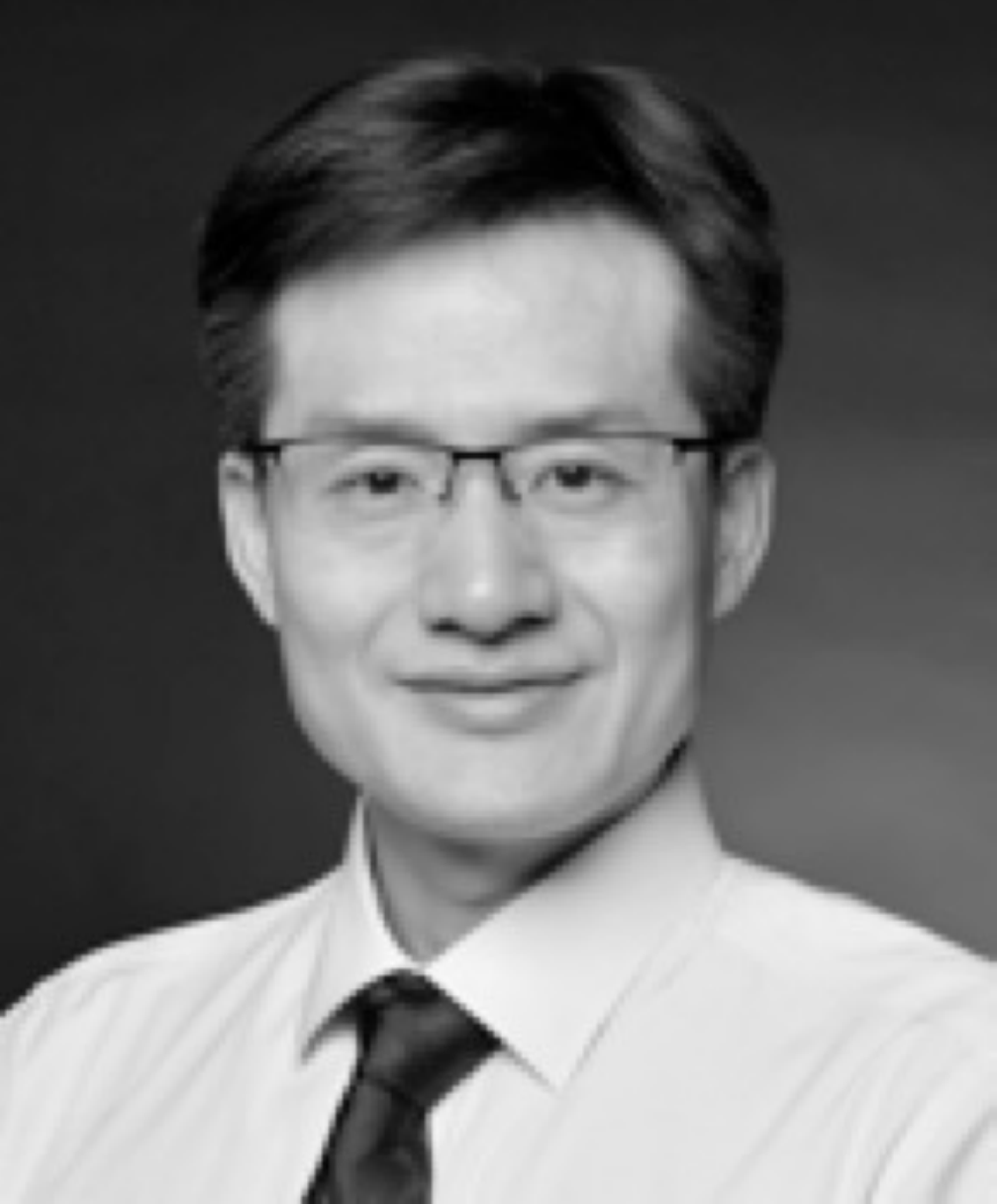}}]{Weiming Wang}
is a Professor with the School of Mechanical Engineering, Shanghai Jiao Tong University, Shanghai, China. His research interests include machine vision, ﬂexible robots, and human-robot interaction.
\end{IEEEbiography}

\begin{IEEEbiography}
    [{\includegraphics[width=1in,height=1.25in,clip,keepaspectratio]{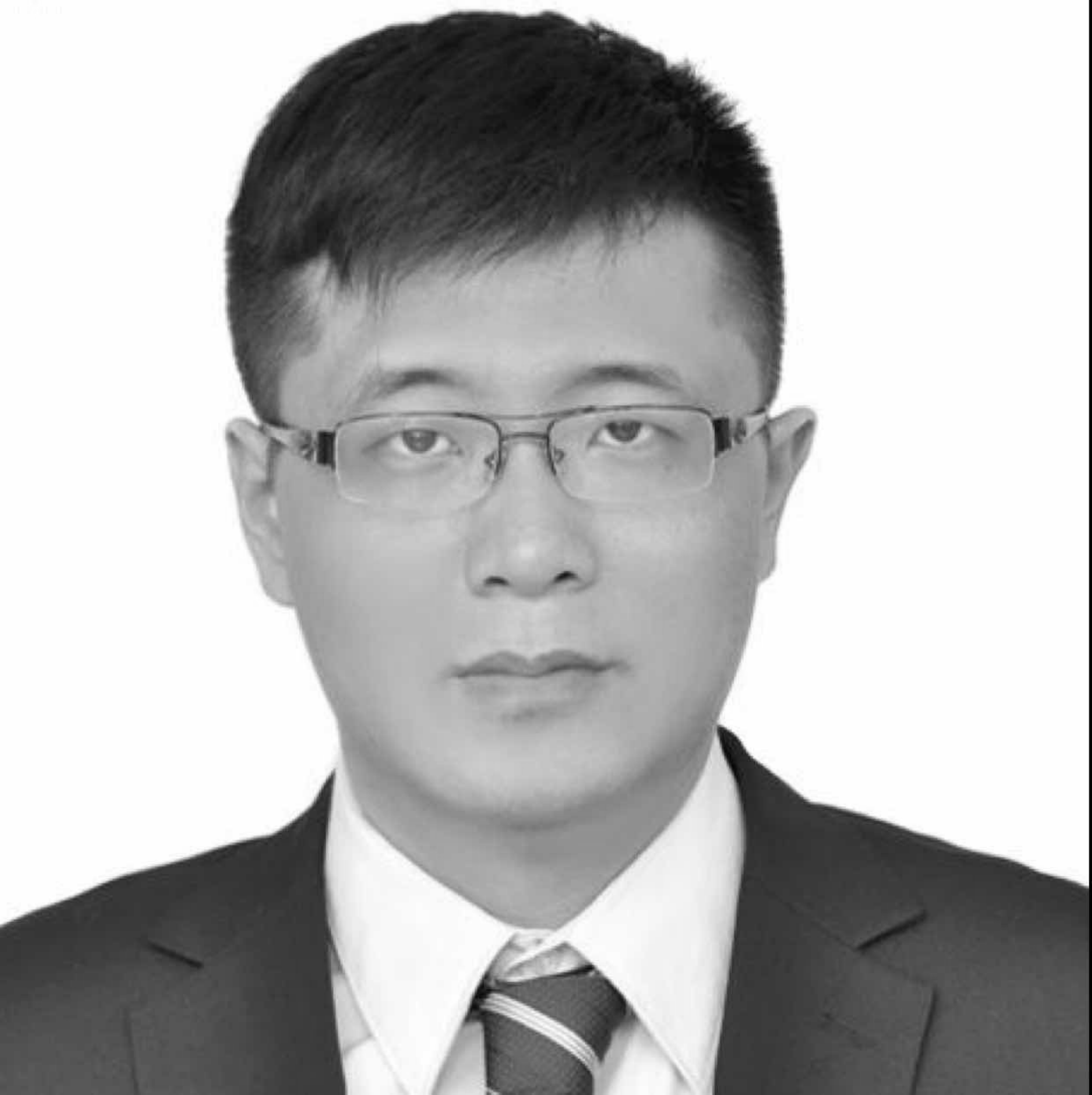}}]{Cewu Lu}
is an Associate Professor at Shanghai Jiao Tong University (SJTU). Before he joined SJTU, he was a research fellow at Stanford University working under Prof. Fei-Fei Li and Prof. Leonidas J. Guibas. He was a Research Assistant Professor at Hong Kong University of Science and Technology with Prof. Chi Keung Tang. He got his PhD degree from the Chinese Univeristy of Hong Kong, supervised by Prof. Jiaya Jia. He is one of the core technique members in Stanford-Toyota autonomous car project. He serves as an associate editor for Journal CVPR and reviewer for Journal TPAMI and IJCV. His research interests fall mainly in computer vision, deep learning, deep reinforcement learning and robotics vision.
\end{IEEEbiography}

\vfill



\end{document}